\documentclass{bmvc2k}

\title{MV-Match: Multi-View Matching for Domain-Adaptive Identification of Plant Nutrient Deficiencies}

\addauthor{Jinhui Yi*}{jinhui.yi@uni-bonn.de}{1}
\addauthor{Yanan Luo*}{yluo@uni-bonn.de}{1}
\addauthor{Marion Deichmann}{madeic@uni-bonn.de}{2}
\addauthor{Gabriel Schaaf}{gabriel.schaaf@uni-bonn.de}{2}
\addauthor{Juergen Gall}{gall@iai.uni-bonn.de}{1,3}
\addinstitution{
 Computer Vision Group\\
 University of Bonn\\
 Bonn, Germany
}
\addinstitution{
 Plant Nutrition Group\\
 University of Bonn\\
 Bonn, Germany
}
\addinstitution{
 Lamarr Institute for Machine Learning and Artificial Intelligence\\
 Germany \\
\vspace{2mm}
\hspace{-3mm} * indicates equal contribution
}

\runninghead{Yi et al.}{MV-Match}

\def\eg{\emph{e.g}\bmvaOneDot}

\def\etal{\emph{et al}\bmvaOneDot}

\usepackage{comment}
\usepackage{tabularx}
\usepackage{booktabs}
\usepackage{multirow}
\usepackage{float}
\usepackage{amssymb}
\usepackage{soul}
\usepackage{xcolor}
\usepackage{subcaption}

\def\ie{\textit{i.e.,~}}
\def\eg{\textit{e.g.,~}}
\def\etal{\textit{et al.~}}
\newcommand{\todo}[1]{{\textbf{\textcolor{blue}{#1}}}}

\begin{document}

\maketitle

\begin{abstract}
An early, non-invasive, and on-site detection of nutrient deficiencies is critical to enable timely actions to prevent major losses of crops caused by lack of nutrients. While acquiring labeled data is very expensive, collecting images from multiple views of a crop is straightforward. Despite its relevance for practical applications, unsupervised domain adaptation where multiple views are available for the labeled source domain as well as the unlabeled target domain is an unexplored research area. In this work, we thus propose an approach that leverages multiple camera views in the source and target domain for unsupervised domain adaptation. We evaluate the proposed approach on two nutrient deficiency datasets. The proposed method achieves state-of-the-art results on both datasets compared to other unsupervised domain adaptation methods. The dataset and source code are available at \href{https://github.com/jh-yi/MV-Match}{https://github.com/jh-yi/MV-Match}. 

\end{abstract}

\section{Introduction}
\label{sec:intro}
In recent years, nutrient status monitoring has become a popular topic for the precision management of fertilizer in smart farming \cite{ali2017leaf, mee2017detecting, padilla2018proximal, rueda2018impact, storm2024research}. An early, non-invasive, and on-site identification of plant nutrient deficiencies is critical to enable timely actions to prevent major losses of crops caused by lack of nutrients. This can help farmers improve crop yields and prevent excess fertilization with negative environmental consequences, such as nitrous oxide emissions or groundwater pollution.  
For rapid and non-invasive diagnosis of nutrient deficiencies, deep learning methods applied to RGB images have been widely adopted \cite{patel2021identification, sudhakar2023computer, talukder2023nutrients, sharma2022ensemble, gul2023exploiting, yi2020deep, tran2019comparative}.
These approaches, however, do not generalize to genotypes that have not been observed during training. Furthermore, collecting labeled data is extremely expensive since it requires expert knowledge to recognize nutrient deficiency and it is even difficult for experts to distinguish plant stress caused by nutrient imbalances from other causes like droughts or diseases. Detecting nutrient deficiencies across different genotypes is thus a highly relevant application for evaluating and developing unsupervised domain adaptation methods \cite{pan2009survey, wang2018deep, tanveer2023depth}. 

Current datasets and approaches for domain adaptation consider only a single view, i.e., each object or scene in the source and target domain have been taken from a single view. Multiple views of the same scene, however, provide additional information since it allows to learn or discard variability of the appearance that is caused by viewpoint changes and not necessarily by changes of the domain. Utilizing this additional information is very practical since it is straightforward for many applications like nutrient deficiency detection to collect multiple views of the same location both for the source and target domain, in particular if the images can be simply taken by a smartphone without the need of any camera calibration or expensive camera setup.          
Despite its practical relevance, unsupervised domain adaptation under multi-view scenarios is a widely unexplored research area. An exception is a recent work by Lu et al.~\cite{lu2023multi} that addresses domain adaptation for object detection in a surveillance setting where the camera views of the target domain overlap. 
This specific setting, however, does not generalize to other tasks where the assumption of overlapping views cannot be guaranteed. 

\begin{figure}[t]
   \begin{center}
   \includegraphics[width=0.5\linewidth]{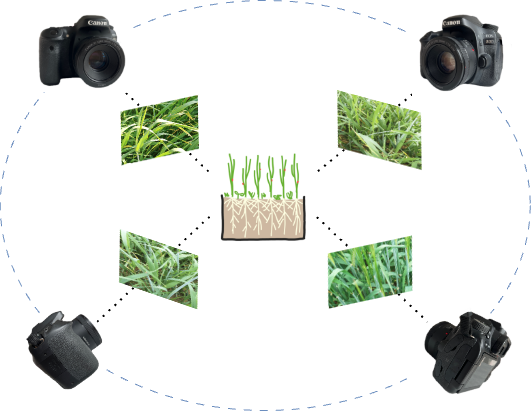}
   \end{center}
      \caption{Illustration of our multi-view setting. Related views are expected to share the same nutrient status, both in the labeled source and the unlabeled target domain. We do not assume that each view contains exactly the same plant, but very closely located plants with the same nutrient status, which makes the data capturing in an open field very simple. 
      }
   \label{fig:teaser}
\end{figure}

In this work, we thus investigate the task of unsupervised multi-view domain adaptation. In our context, multi-view refers to multiple images of the same scene taken from different views, which is different to the multi-view learning approach investigated in Xia et al.~\cite{xia2022incomplete} where views refer to different modalities. To study this challenging task, we collected a dataset of images of crops that suffer from different nutrient deficiencies. Each crop has been captured over the growing season multiple times and each time from multiple views as illustrated in Figure~\ref{fig:teaser}. While each view does not necessarily contain exactly the same plant, the multiple views contain very closely located plants with the same nutrient status.
The images are annotated by the date, genotype, and nutrient deficiency. As a second contribution, we propose an approach that leverages the multiple views for domain adaptation. The proposed approach, which we term Multi-View Match (MV-Match), enforces the consistency of the predictions among multiple views. In addition, we propose a \textbf{S}imilarity-\textbf{g}uided \textbf{V}iew \textbf{M}ining (SgVM) mechanism to automatically select the most dissimilar views that contain complementary information given a query image.  

We evaluate our approach on two nutrient deficiency benchmarks and show that our approach achieves state-of-the-art performance compared to other unsupervised domain adaptation approaches. 

\section{Related Work} 
\label{sec:related}

\paragraph{Plant Nutrient Deficiency.} 
Better matching of the timing and amount of fertilizer inputs to plant requirements will improve nutrient use efficiency and crop yields. 
Driven by tremendous economic potential, deep learning applied to RGB images is widely adopted for early, non-invasive, and on-site monitoring of plant nutrient deficiency in plants \cite{sudhakar2023computer}, including rice \cite{talukder2023nutrients, sharma2022ensemble}, sugar beet \cite{yi2020deep}, hydroponic basil \cite{gul2023exploiting}, winter wheat \cite{yi4549653non}, and tomato \cite{tran2019comparative}. 
These approaches evaluated various convolutional neural networks (CNNs) on their proposed datasets \cite{yi2020deep, gul2023exploiting, yi4549653non, tran2019comparative} or applied an ensemble of CNNs to achieve better performance on publicly available datasets \cite{talukder2023nutrients, sharma2022ensemble}. 
However, they detected plant nutrient deficiency in images of the same domain as the labeled training data, whereas we aim to identify the nutrient status of plants in out-of-domain data. 

\paragraph{Unsupervised Domain Adaptation.} 
Unsupervised domain adaptation approaches \cite{pan2009survey, wang2018deep, tanveer2023depth} aim to adapt a model trained on a labeled source domain to an unlabeled target domain. 
Due to ubiquitous domain shifts in real life, unsupervised domain adaptation approaches were proposed for various applications, including image classification \cite{long2015learning, ganin2016domain, chen2019transferability}, semantic segmentation \cite{vu2019advent, araslanov2021self, hoyer2023mic}, and smart farming \cite{schmitter2017unsupervised, gogoll2020unsupervised, rehman2022deep, ma2024transfer, abdalla2024assessing}. Most of these approaches involve the minimization of discrepancies, adversarial training, or self-training. Methods with discrepancy minimization aim to minimize the domain gap by adopting a statistical distance function such as maximum mean discrepancy \cite{long2015learning, long2017deep}, correlation alignment \cite{sun2016return, sun2016deep}, or entropy minimization \cite{grandvalet2004semi, vu2019advent}. 
In adversarial training, encoders and domain discriminators are trained to learn domain-invariant inputs \cite{gong2019dlow, hoffman2018cycada}, features \cite{ganin2016domain, long2018conditional}, or outputs \cite{saito2018maximum, tsai2018learning, vu2019advent}. Recently, unsupervised domain adaptation approaches using pseudo-labels were proposed to generate artificial supervised signals based on confidence thresholds \cite{mei2020instance, zou2018unsupervised} or pseudo-label prototypes \cite{pan2019transferrable, zhang2021prototypical} for the unlabeled target data.  
Although unsupervised domain adaptation for smart farming \cite{ma2024transfer} is attracting more attention for species recognition \cite{gogoll2020unsupervised}, disease detection \cite{abdalla2024assessing}, drought stress \cite{schmitter2017unsupervised}, and relative water content prediction \cite{rehman2022deep}, the exploration of unsupervised domain adaptation for identification of plant nutrient deficiency \cite{yi2020deep} remains an open area of investigation. Moreover, leveraging multiple views of a plant or object in the source and target domain is an unexplored research area in the field of domain adaptation.

\paragraph{Multi-View Learning} Multi-view learning is commonly adopted in 3D reconstruction \cite{wu2024multi}, 3D object recognition \cite{qi2021review}, and perception of 2D images. For 2D images, multi-view consistency is a form of consistency regularization in semi-supervised learning, which is often applied to ensure consistency over different data augmentations \cite{araslanov2021self, choi2019self, melas2021pixmatch} and different crops \cite{hoyer2022hrda, lai2021semi}. 
However, the aforementioned approaches for 2D images only consider multi-views as augmented versions of the same image, which inevitably contain local patterns instead of the holistic appearance of a given object. 
In this work, we make use of images of plants with various camera angles to include different light conditions, viewpoints, and crop parts, to incorporate holistic details of diagnosed plants. We enforce consistency among these multi-view images instead of augmented images. 
Concurrent works \cite{lu2023multi, xia2022incomplete} evaluated similar settings for object detection, image classification, and 3D reconstruction, but none of them can be applied to the task of nutrient deficiency detection. 


\begin{figure}[t]
   \begin{center}
   \includegraphics[width=\textwidth]{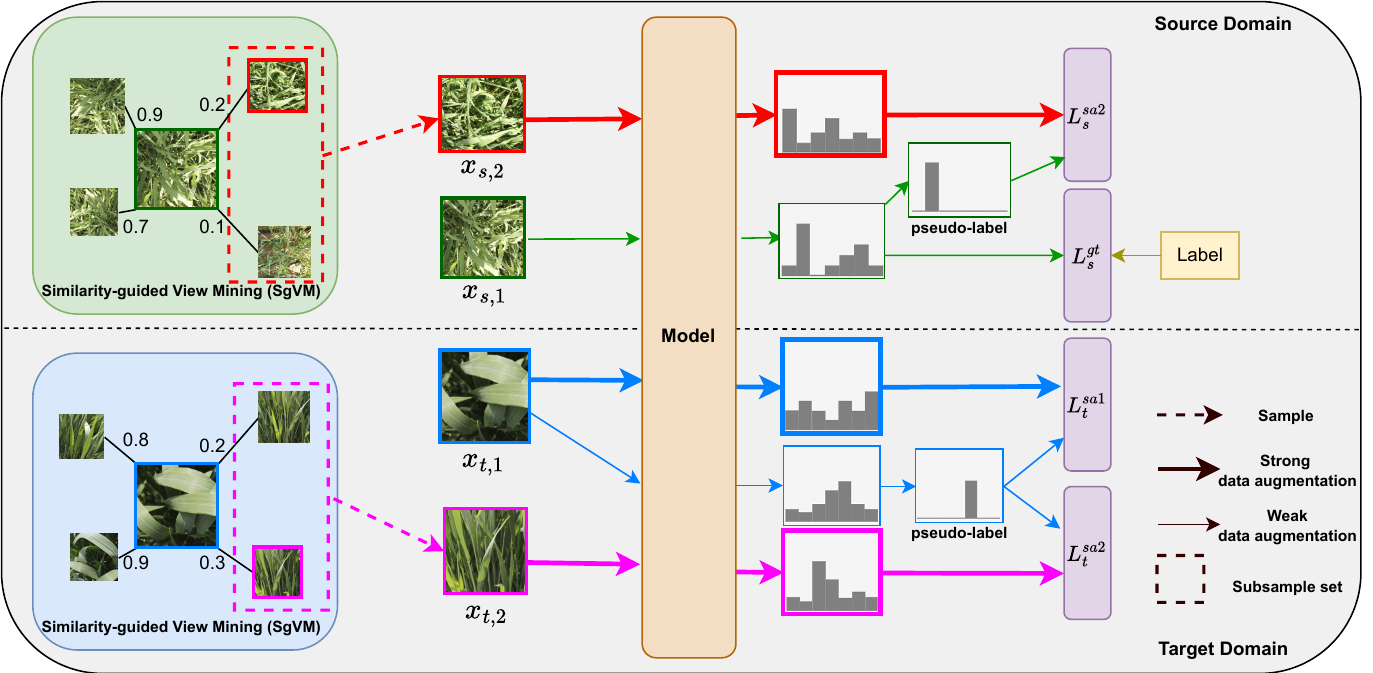}
   \end{center}
      \caption{Proposed approach for unsupervised domain adaptation. Given multiple views of a crop in the labeled source domain (top) and the unlabeled target domain (bottom), a random query image is sampled from the source and the target domain (center of the green and blue box). The \textbf{S}imilarity-\textbf{g}uided \textbf{V}iew \textbf{M}ining (SgVM) module then computes the normalized mutual information between each query-view pair to select the top n
      dissimilar views of the same crop (red dashed rectangles). From these two sets, we randomly select a second image for each query image. We then apply weak or strong data augmentation to the four images, i.e., the two query images and their corresponding view pair images, and feed them to a shared model for predicting nutrient deficiencies. While the prediction of the query image of the source domain is supervised by the ground-truth label, the other predictions are enforced to be consistent with the corresponding view pair. For this, the query image with weak augmentation is always considered as a reference prediction, both for the source and target domain.    
      }
   \label{fig:framework}
\end{figure}

\section{MV-Match} 
\label{sec:method}
In this work, we propose an approach for unsupervised multi-view domain adaptation. This means that we have multiple views both in the source and the training domain, as illustrated in Fig.~\ref{fig:teaser}. Such a scenario is highly relevant for applications since capturing multiple views of the same crop or object is straightforward whereas annotating images by trained experts is often very expensive.  

Our approach, which is illustrated in Fig.~\ref{fig:framework}, thus aims to leverage the multiple views in the source and target domain. While we have labeled images for the source domain, where $x_{s, i}$ denotes an image of the source domain and $y_{s, i} \in \{1, 2, \cdots, C\}$ is the corresponding label, the images $x_{t, i}$ of the target domain are unlabeled. In our case, the images of the source and target domain contain different genotypes and the goal is to recognize different types of nutrient deficiencies in the target domain.       

As illustrated in Fig.~\ref{fig:framework}, we train a joint model for the source and target domain and aim to learn a model that is agnostic to the domain. During training, we select for each batch the same number of query images from the source and target domain. For each query image, we sample a second image that shows the same crop but from a different view. This step will be described in Section~\ref{sec:sgvm}. For a query image of the source domain $x_{s,i}$, we can directly apply the cross-entropy loss using the ground-truth label $y_{s,i}$:  
\begin{equation}
   \mathcal{L}_s^{gt} (x_{s,i}, y_{s,i}) = H(y_{s,i}, P_{x_{s,i}^{wa}}) = - \sum_{c=1}^{C} \mathbf{1}(c=y_{s,i}) \log P_{x_{s,i}^{wa}}(c),
   \label{eq:s}
\end{equation}
where $P_x$ is the prediction of the network for image $x$ and $P_x(c)$ is the predicted probability for class $c$. The indicator function $\mathbf{1}(c=y_{s,i})$ is 1 for the ground-truth label and 0 otherwise. $x^{wa}$ is the weakly augmented version of a given image $x$, which is a random horizontal flip with a probability of 50\%. 

For the other images, we introduce consistency loss functions, which are denoted by $\mathcal{L}_s^{sa2}$, $\mathcal{L}_t^{sa1}$, and $\mathcal{L}_t^{sa2}$. These loss functions measure the prediction consistency across views for the source and the target domain, as illustrated in Fig.~\ref{fig:framework}. We describe them in the next section more in detail. 

\subsection{Multi View Matching} 
For each query image of the source or target domain, we have a set of images $\mathcal{M}_{s,i}$ or $\mathcal{M}_{t,i}$ that show the same crop but from a different view. In Section~\ref{sec:sgvm}, we will describe how these two sets are obtained. For each query image $x_{s,i}$ and $x_{t,i}$, we randomly select a second image from the corresponding set $x_{s,j} \in \mathcal{M}_{s,i}$ or $x_{t,j} \in \mathcal{M}_{t,i}$. As illustrated in Fig.~\ref{fig:framework}, we apply strong augmentation to the sampled views, which we denote by $x_{s,j}^{sa}$ and $x_{t,j}^{sa}$. For this, we adopt AutoAugment \cite{cubuk2019autoaugment} as strong augmentation, which is described more in detail in the supplementary material. For sampled views of the source domain, we compute the cross-entropy to enforce that the prediction is the same as for the query image of the source domain:      
\begin{align}
   \mathcal{L}_s^{sa2} (x_{s,i}, x_{s,j}) = \mathbf{1}(\max_c P_{x_{s,i}^{wa}}(c) \geq \tau )H(\hat{y}_{s,i}, P_{x_{s,j}^{sa}}).
\label{eq:loss_sa2}
\end{align}
For computing the cross-entropy $H$, we consider in the experiments two cases: a) $\hat{y}_{s,i} = \operatorname{argmax}_c P_{x_{s,i}^{wa}}(c)$ or b) $\hat{y}_{s,i} = P_{x_{s,i}^{wa}}$. While a) generates a hard pseudo-label, b) is a soft labeling based on the predicted probabilities of all classes. In the case of a hard pseudo-label, the loss is only applied if the prediction on the query image reaches a certain confidence, i.e., $\max_c P_{x_{s,i}^{wa}}(c) \geq \tau$.             

For the target images, we define a similar loss function: 
\begin{align}
   \mathcal{L}_t^{sa2} (x_{t,i}, x_{t,j}) = \mathbf{1}(\max_c P_{x_{t,i}^{wa}}(c) \geq \tau )H(\hat{y}_{t,i}, P_{x_{t,j}^{sa}}),
\label{eq:loss_ta2}
\end{align}
which enforces that the predictions of the two views of the target domain are consistent. In addition, we compute a loss that measures the consistency of the weak and strong augmentation of the target query image:  
\begin{align}
   \mathcal{L}_t^{sa1}(x_{t,i}) = \mathbf{1}(\max_c P_{x_{t,i}^{wa}}(c) \geq \tau )H(\hat{y}_{t,i}, P_{x_{t,i}^{sa}}).
\label{eq:loss_ta1}
\end{align}

For training, we combine all four loss terms without any weighting:  
\begin{align}
\mathcal{L} = \mathcal{L}_s^{gt} (x_{s,i}, y_{s,i}) + \mathcal{L}_s^{sa2} (x_{s,i}, x_{s,j}) + \mathcal{L}_t^{sa2} (x_{t,i}, x_{t,j}) + \mathcal{L}_t^{sa1} (x_{t,i})
\end{align}
The impact of the loss functions except $\mathcal{L}_s^{gt}$, which is always needed, is evaluated in the experiments.

\subsection{Similarity-guided View Mining}
\label{sec:sgvm}

For a given query image $x_{s,i}$ or $x_{t,i}$, we need to select the set of images $\mathcal{M}_{s,i}$ or $\mathcal{M}_{t,i}$. We first select all images that have been taken from the same location as the query image but from a different view. Since the nutrient status of the same plant might change over time, we only consider images that have been taken on the same day as the query image.    

In the experiments, we show that it is better to first select a subset of images instead of randomly sampling $x_{s,j}$ from $\mathcal{M}_{s,i}$. To this end, we select a subset of images that are most dissimilar to the query image, but that show the same crop as the query image. In this way, we select views that are not very similar to the query view.             

For the selection, we compute the Normalized Mutual Information (NMI) \cite{vinh2010information} between the query image $x_{s,i}$ and all view images $x_{s,k}$ of the same crop as the query image:  
\begin{align}
NMI(x_{s,i}, x_{s,k}) = 2\frac{H(x_{s,i}) - H(x_{s,i} | x_{s,k})}{H(x_{s,i})+H(x_{s,k})}.
\end{align}
where $H(x_{s,i})$ is the entropy of image $x_{s,i}$ and $H(x_{s,i}|x_{s,k})$ is the conditional entropy. 
Larger $NMI(x_{s,i}, x_{s,k})$ indicates higher similarity. For building $\mathcal{M}_{s,i}$, we select the 5 images $x_{s,k}$ with the lowest NMI, i.e., we select the 5 most dissimilar images to the query image. We denote the selection as Similarity-guided View Mining (SgVM). We apply SgVM to the target images in the same way to obtain $\mathcal{M}_{t,i}$.

\section{Experiments}
\label{sec:exp}

\subsection{Datasets and Metrics}
We evaluate our proposed MV-Match for image classification on two nutrient deficiency benchmarks, MiPlo-B \cite{deichmann2024rgb} and MiPlo-WW. 
We report the top-1 accuracy on the test set as our metric. We also report per-nutrient accuracy as it is critical for nutrient monitoring in smart farming. As a network, we use a ResNet50 \cite{he2016deep}. 

\textbf{MiPlo-B.} 
The Mini Plot Barley (MiPlo-B) dataset consists of  18559 images with 6 nutrient treatments (-N, -P, -K, -B, -S, ctrl) annotated, ranging from 21.06.2022 - 20.07.2022 (16 dates). It contains two genotypes: Barke (9305 images) and Hanna (9254 images). For each genotype, each treatment was repeated 4 times, resulting in 24 containers, each of which has a unique ID. Six unique containers with six different nutrient treatments were selected as the test set while the other containers as the training set (\#train:\#test$\approx$75\%:25\%), trying to avoid information leaks due to commonly adopted random sampling, i.e., multi-views of the same crop being separated into both training and test set. 

\textbf{MiPlo-WW.} The Mini Plot Winter Wheat (MiPlo-WW) dataset has 12466 images with 6 treatments (-N, -P, -K, -B, -S, ctrl) annotated, ranging from 12.05.2023 - 24.05.2023 (13 dates). It contains two genotypes: Julius (6253 images) and Meister (6213 images). The ID settings are the same as above. 

 We refer to the supplementary materials for more details about the datasets and experimental details. 

\begin{table}[t] 
   \caption[SOTA B2H]
      {Top-1 Classification Accuracy (\%) for adaptation across genotypes: \textbf{Barley: Barke $\rightarrow$ Hanna}. \textit{Oracle} indicates the model was trained with full supervision on the \textbf{Hanna} training set. \textit{Source-Only} denotes the results without adaptation. 
      The highest accuracy is shown in bold, while the second best is underlined. 
      \textit{TPS} refers to throughput per second. 
      }
      \begin{center}
         \resizebox{0.95\textwidth}{!}{
         \begin{tabular}{cccccccccc|c|cc} 
            \toprule  
            \multicolumn{4}{c}{\multirow{2}{*}{\textbf{Model}}}  &  \multicolumn{7}{c}{\textbf{Barley: Barke $\rightarrow$ Hanna}} \\ 
            \cmidrule{5-13}  
            \multicolumn{4}{c}{}   &  \textbf{-N}  &  \textbf{-P}  &  \textbf{-K} &  \textbf{-B} &  \textbf{-S} &  \textbf{control}  &  \textbf{AVG}  &  \textbf{Train Time (min)} &  \textbf{TPS (test)} \\          
            \hline
                  \multicolumn{4}{c}{Oracle} & 97.3 & 78.5 & 82.1 & 99.1 & 9.1 & 87.1 & 75.4 & - & - \\ 
                  \hline
                  \multicolumn{4}{c}{Source-Only} & 95.1 & 26.0 & 80.4 & 84.3 & 14.8 & 24.2 & 54.0  & - & - \\ 
                   \multicolumn{4}{c}{DANN \cite{ganin2016domain}}  &  93.0   &  36.6   &  59.0  & 68.6   & 24.2  & 46.0  & 54.7 & 215.1 & 35.0 \\ 
                   \multicolumn{4}{c}{ADDA \cite{tzeng2017adversarial}}  &  83.7   &  47.1   &  50.4  & 79.5   & 17.4  & 33.0  & 51.8 & 125.4 & 35.0 \\ 
                   \multicolumn{4}{c}{JAN \cite{long2017deep}}  &   87.2  &  26.1   &  63.7  &  70.8  & 22.9  & 39.6  &  51.9  & 86.6 & 35.0 \\ 
                   \multicolumn{4}{c}{CDAN \cite{long2018conditional}}  &  95.2   &  35.0   &  41.5  & 69.8   & 20.4  & 44.8  & 51.3 & 100.3 & 35.0  \\ 
                   \multicolumn{4}{c}{BSP \cite{chen2019transferability}}  &  94.5   &  41.3   &  68.2  & 76.0   & 14.5  & 57.2  & 58.8 & 153.0 & 35.0 \\ 
                   \multicolumn{4}{c}{AFN \cite{xu2019larger}}  &  94.7   &  25.8   &  73.9  & 64.1   & 24.9  & 68.7  & 59.0 & 233.9 & 35.0 \\ 
                   \multicolumn{4}{c}{Mean Teacher \cite{tarvainen2017mean}} & 94.2 & 32.1 & 81.8 & 89.4 & 11.7 & 52.5 & 60.6 & 266.7 & 35.0 \\ 
                   \multicolumn{4}{c}{FixMatch \cite{sohn2020fixmatch}}  & 98.2 & 38.9 & 83.8 & 86.4 & 25.4 & 53.2 & 64.6 & 257.6 & 35.0 \\ 
                  \multicolumn{4}{c}{FlexMatch \cite{zhang2021flexmatch}} & 97.2 & 50.0 & 89.1 & 89.4 & 11.0 & 58.0 & 65.9 & 276.4 & 35.0  \\ 
                  \multicolumn{4}{c}{Ours+hard} & 92.7 & 50.0 & 86.8 & 95.3 & 22.7 & 68.9 & \textbf{69.6} & 666.7 & 35.0 \\ 
                   \multicolumn{4}{c}{Ours+soft} & 97.0 & 53.9 & 75.4 & 92.8 & 20.0 & 64.7 & \underline{67.4} & 714.3 & 35.0 \\ 
                   \midrule         
         \end{tabular} }
      \end{center}
      \label{tab:B2H}
      \vspace{-5mm}
\end{table}

\begin{table}[t] 
   \caption[SOTA J2M]{Top-1 Classification Accuracy (\%) for adaptation across genotypes: \textbf{Winter Wheat: Julius $\rightarrow$ Meister}. \textit{Oracle} indicates the model was trained with full supervision on the \textbf{Meister} training set. \textit{Source-Only} denotes the results without adaptation. 
   The highest accuracy is shown in bold, while the second best is underlined. 
   \textit{TPS} refers to throughput per second. 
   }
      \begin{center}
         \resizebox{0.95\textwidth}{!}{
         \begin{tabular}{cccccccccc|c|cc} 
            \toprule  
            \multicolumn{4}{c}{\multirow{2}{*}{\textbf{Model}}}  &  \multicolumn{7}{c}{\textbf{Winter Wheat: Julius $\rightarrow$ Meister}} \\ 
            \cmidrule{5-13}  
            \multicolumn{4}{c}{}   &  \textbf{-N}  &  \textbf{-P}  &  \textbf{-K} &  \textbf{-B} &  \textbf{-S} &  \textbf{control}  &  \textbf{AVG}  &  \textbf{Train Time (min)} &  \textbf{TPS (test)}  \\    
            \hline
                   \multicolumn{4}{c}{Oracle}  & 98.9 & 85.4 & 58.9 & 81.9 & 50.2 & 51.7 & 71.0  & - & - \\ 
                   \hline
                   \multicolumn{4}{c}{Source-Only}  & 100.0 & 88.8 & 5.6 & 46.6 & 1.1 & 61.6 & 50.5  & - & - \\ 
                   \multicolumn{4}{c}{DANN \cite{ganin2016domain}} & 99.8 & 54.9 & 22.6 & 52.9 & 20.5 & 58.9 & 51.4 & 196.7 & 35.0 \\  
                   \multicolumn{4}{c}{ADDA \cite{tzeng2017adversarial}} & 99.0 & 57.5 & 50.7 & 51.3 & 30.4 & 44.1 & 55.7 & 130.3 & 35.0 \\ 
                   \multicolumn{4}{c}{JAN \cite{long2017deep}} & 98.9 & 50.0 & 26.7 & 49.6 & 39.9 & 58.2 & 53.9 & 83.3 & 35.0\\  
                   \multicolumn{4}{c}{CDAN \cite{long2018conditional}} & 99.2 & 44.0 & 33.7 & 55.9 & 29.7 & 63.9 & 54.3 & 95.2 & 35.0\\  
                   \multicolumn{4}{c}{BSP \cite{chen2019transferability}} & 100.0  & 59.7 & 56.3 & 55.9 & 49.8 & 49.4 & 61.9 & 144.4 & 35.0 \\  
                   \multicolumn{4}{c}{AFN \cite{xu2019larger}}  & 99.6 & 67.9 & 24.8 & 68.1 & 19.8 & 61.6 & 56.6 & 199.0 & 35.0 \\ 
                   \multicolumn{4}{c}{Mean Teacher \cite{tarvainen2017mean}} & 99.2 & 67.9 & 62.6 & 72.7 & 18.6 & 37.6 & 59.6 & 266.7 & 35.0 \\ 
                   \multicolumn{4}{c}{FixMatch \cite{sohn2020fixmatch}}  & 99.6 & 75.0 & 66.7 & 62.2 & 25.5 & 64.6 & \underline{65.6} & 279.1 & 35.0 \\ 
                  \multicolumn{4}{c}{FlexMatch \cite{zhang2021flexmatch}} & 99.2 & 72.8 & 68.1 & 87.4 & 3.0 & 64.6 & 65.5  & 239.3 & 35.0  \\ 
                   \multicolumn{4}{c}{Ours+hard} & 99.2 & 59.0 & 71.1 & 93.7 & 4.6 & 76.4 & \textbf{66.9} & 588.2 & 35.0 \\ 
                   \multicolumn{4}{c}{Ours+soft} & 100.0 & 60.8 & 59.3 & 76.5 & 29.7 & 50.6 & 62.5 & 555.6 & 35.0 \\ 
                   \midrule         
         \end{tabular} }
      \end{center}
      \label{tab:J2M}
      \vspace{-5mm}
\end{table}

\subsection{Comparison to State of the Art}

We present experimental results of our approach compared to different baselines. \textit{Source-only} denotes training only on the source domain without adaptation. 
DANN \cite{ganin2016domain}, ADDA \cite{tzeng2017adversarial}, JAN \cite{long2017deep}, and CDAN \cite{long2018conditional} learn domain-invariant features with adversarial learning: DANN jointly trains encoder and discriminator with a gradient reversal layer; ADDA adopts separated encoders for source and target domains and alternates training of the encoder and discriminator; JAN adopts adversarial training to maximize a joint maximum mean discrepancy (JMMD);  CDAN conditions the adversarial adaptation models on both the features and the classifier predictions.  BSP \cite{chen2019transferability} avoids deterioration of the feature discriminability after adaptation by penalizing the largest singular values. AFN \cite{xu2019larger} enlarges feature norms to enhance the transferability of features. Mean Teacher \cite{tarvainen2017mean} constructs a teacher model by averaging model weights of a student model with an exponential moving average and forcing the consistency between the predictions of two models. FixMatch \cite{sohn2020fixmatch} takes the predictions of weakly-augmented unlabeled images as pseudo-labels and their strongly-augmented version as predictions to train the model in a supervised manner. FlexMatch \cite{zhang2021flexmatch} extends FixMatch by considering adaptive thresholds for different classes at each time step. All approaches were trained with the same amount of images, i.e., single-view approaches were trained using all views as images but they do not utilize multi-view information.   
In what follows, we show the performance of our model in two plant nutrient deficiency benchmarks under different settings, \ie adaptation across genotypes (smaller domain gap) and across cultivars (larger domain gap).

\textbf{Adaptation Across Genotypes: } We report in Tables \ref{tab:B2H} and \ref{tab:J2M} the performance of adaptation across genotypes in terms of per-nutrient accuracy and average accuracy. 
The results indicate that for both settings of adaptation across genotypes, MV-Match outperforms the baseline methods by a large margin. 
Comparing \textit{Oracle} and \textit{Source-Only}, we see that different genotypes drastically deteriorate the performance without adaptation, especially for \textit{P} deficiency and \textit{control} in \textbf{Barley: Hanna $\rightarrow$ Barke} and \textit{K} deficiency, \textit{B} deficiency and \textit{S} deficiency in \textbf{Winter Wheat: Meister $\rightarrow$ Julius}. 
On adapting the network, most baselines based on adversarial learning or discrepancy minimization do not attain large improvements over the un-adapted model for both scenarios. Notably, Mean Teacher and FixMatch achieve a relatively large improvement. Both approaches adopt consistency regularization, which we assume is critical for plant nutrient deficiency. With the proposed multi-view consistency, our approach yields consistent and significant performance gains in both adaptation settings. MV-Match with hard pseudo-labels outperforms its counterpart with soft pseudo-labels in both settings.  
We furthermore report the training time and throughput per second (TPS) during inference in Tables \ref{tab:B2H} and \ref{tab:J2M}. While the inference time is the same for all methods since they all use the same backbone suitable for real-time processing, our approach requires more time for training due to selecting dissimilar views.


\textbf{Adaptation Across Cultivars: }
We also evaluate the methods for adaptation across different crop species. This setting is extremely challenging because both crops were cultivated in different years, where changing environmental conditions as well as abiotic and biotic stresses might have affected the growing status of crops. This is reflected in the results in Table \ref{tab:cross}. 
For many adaptation settings, baseline methods decrease the performance of the un-adapted model. Our method consistently boosts the performance across crop species. We attribute this robustness against the changing environment to our proposed multi-view matching, which incorporates holistic visual symptoms for effective identification of plant nutrient deficiency. 
Our model with soft pseudo-labels works better than with hard pseudo-labels. This is different to the across genotypes experiments where the domain gaps are smaller. This is not unexpected since soft pseudo-labels work better for out-of-domain unlabeled data \cite{xie2020self}.
Note that FixMatch and FlexMatch also achieve promising results in adaptation across genotypes, but their performance deteriorates for adaptation across crop species.

\begin{table}[t] 
   \caption[SOTA cross]{Mean Top-1 Classification Accuracy (\%) for adaptation across crop species.
   \textit{Oracle} indicates the model was trained with full supervision on the target training set. \textit{Source-Only} denotes the results without adaptation. 
   The highest accuracy is shown in bold, while the second best is underlined. 
   }  
      \begin{center}
         \resizebox{0.9\textwidth}{!}{
         \begin{tabular}{cccccccc|cccc} 
            \toprule  
            \multicolumn{4}{c}{\multirow{2}{*}{\textbf{Model}}}  & \multicolumn{4}{c}{Barley $\rightarrow$ Winter Wheat} & \multicolumn{4}{c}{Winter Wheat $\rightarrow$ Barley}\\ 
            \cmidrule{5-12}  
            \multicolumn{4}{c}{}  & B $\rightarrow$ J & B $\rightarrow$ M & H $\rightarrow$ J & H $\rightarrow$ M & J $\rightarrow$ B & J $\rightarrow$ H & M $\rightarrow$ B & M $\rightarrow$ H \\ 
            \hline
                   \multicolumn{4}{c}{Oracle} & 73.5 & 71.0 & 73.5 & 71.0 & 67.9 & 75.4 & 67.9 & 75.4 \\ 
                   \hline
                   \multicolumn{4}{c}{Source-Only} & 31.9 & 40.7 & 27.0 & 31.7 & 18.8 & 13.3 & 16.3 & 13.7 \\ 
                   \multicolumn{4}{c}{DANN \cite{ganin2016domain}} & \underline{40.5} & 39.4 & 22.2 & 19.3 & 19.0 & 18.7 & 23.8 & 17.6 \\ 
                   \multicolumn{4}{c}{ADDA \cite{tzeng2017adversarial}} & 30.6 & 34.4 & 21.3 & 24.4 & \underline{25.2} & 16.0 & 21.0 & 19.8\\ 
                   \multicolumn{4}{c}{JAN \cite{long2017deep}} & 35.5 & 40.0 & 14.4 & 27.4 & 19.5 & 14.6 & 17.9 & 16.1\\ 
                   \multicolumn{4}{c}{CDAN \cite{long2018conditional}} & 36.0 & 34.3 & 25.7 & 17.2 & 21.9 & 14.0 & 24.4 & 18.0 \\ 
                   \multicolumn{4}{c}{BSP \cite{chen2019transferability}}  & 39.9 & \underline{42.5} & 30.5 & 29.6 & 22.1 & 16.3 & 21.0 & 18.3 \\ 
                   \multicolumn{4}{c}{AFN \cite{xu2019larger}} & 38.2 & 39.6 & 23.3 & 20.0 & 20.9 & 18.0 & 22.2 & 18.5 \\ 
                   \multicolumn{4}{c}{Mean Teacher \cite{tarvainen2017mean}}  & 30.6 & 37.6 & \textbf{37.1} & \underline{40.9} & 23.3 & \underline{19.9} & \textbf{26.0} & \underline{20.4} \\ 
                   \multicolumn{4}{c}{FixMatch \cite{sohn2020fixmatch}}  & 27.9 & 33.6 & 29.4 & 27.8 & 24.3 & 19.3 & 19.6 & 19.5 \\ 
                  \multicolumn{4}{c}{FlexMatch \cite{zhang2021flexmatch}} & 26.8 & 35.4 & 30.1 & 34.3 & 25.1 & 16.5 & 21.7 & 18.9 \\ 
                   \multicolumn{4}{c}{Ours+hard}  & 33.7 & 33.7 & 28.9 & 31.8 & 20.9 & 16.8 & 21.6 & 16.5 \\ 
                   \multicolumn{4}{c}{Ours+soft}  & \textbf{42.2} & \textbf{46.6} & \underline{34.4} & \textbf{41.2} & \textbf{27.1} & \textbf{23.2} & \textbf{26.0} & \textbf{21.3} \\ 
                   \midrule         
         \end{tabular}}
      \end{center}
      \label{tab:cross}
      \vspace{-5mm}
\end{table}

\subsection{Ablation Study}
The results in Table \ref{tab:mv} show that forcing consistency between an unlabeled target image with its related view significantly boosts the performance, and incorporating such consistency for the source image brings extra gain in performance. This validates the effectiveness of our proposed multi-view matching mechanism. Note that we only use weak augmentation instead of strong augmentation for these results. To further explore the effects of strong augmentation in our multi-view setting, we run a series of experiments with various combinations of loss components, as shown in Table \ref{tab:strong}. 
The models that use $L_s^{sa2}$ and $L_t^{sa2}$ have better results as compared to Table \ref{tab:mv}, indicating the effectiveness of strong augmentation. By including $L_t^{sa1}$, we observe a decent boost in performance. The combination of $L_s^{sa2}$, $L_t^{sa2}$ and $L_t^{sa1}$ achieves the highest accuracy of 69.6\% on the Barley: B $\rightarrow$ H benchmark. Such a result indicates that complementary information is learned by multi-view matching. More ablation studies are in the supplementary material.


\vspace{4mm}
\begin{minipage}[c]{0.45\textwidth}
   \begin{center}
   \begin{tabular}{|cc|c|}
      \hline
       $\mathcal{L}_s^{wa2}$ & $\mathcal{L}_t^{wa2}$ & B $\rightarrow$ H \\ 
      \hline
       & & 54.0 \\
       & \checkmark & 65.5 \\
       \checkmark & \checkmark & \textbf{66.0} \\ 
      \hline
   \end{tabular}
   \end{center}
   \captionof{table}{Ablation study on effects of multi-view matching without any strong augmentation. $\mathcal{L}_s^{wa2}$ denotes the self-training loss between weakly-augmented source image (pseudo-labels) and its weakly-augmented related view (predictions). The same applies to $\mathcal{L}_t^{wa2}$. 
   }
   \label{tab:mv}
\end{minipage}
\hspace{4pt}
\begin{minipage}[c]{0.45\textwidth}
   \begin{center}
   \resizebox{0.8\textwidth}{!}{
   \begin{tabular}{|ccc|c|}
      \hline
       $\mathcal{L}_s^{sa2}$ & $\mathcal{L}_t^{sa2}$ & $\mathcal{L}_t^{sa1}$ & B $\rightarrow$ H\\ 
      \hline
       & & & 54.0 \\
       & \checkmark & & 67.2 \\ 
       \checkmark & \checkmark & & 68.7 \\
      \hline
        &  & \checkmark & 64.6 \\ 
        & \checkmark & \checkmark & 67.3 \\ 
       \checkmark & \checkmark & \checkmark & \textbf{69.6} \\
      \hline
   \end{tabular}}
   \end{center}
   \captionof{table}{Ablation study on the impact of loss functions in the Barley: B $\rightarrow$ H benchmark. 
   }
   \label{tab:strong}
   \vspace{5mm}
\end{minipage}

\subsection{Visualization}

\begin{figure}[tb]
   \begin{center}

   \subfigure[]{
      \includegraphics[width=0.48\textwidth]{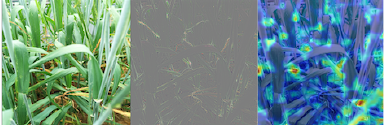} 
   }
   \subfigure[]{
      \includegraphics[width=0.48\textwidth]{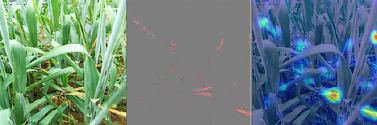}
   }
   \end{center}
      \caption{Saliency visualization (a) without adaptation and (b) with adaptation. For each case, we show the original image, saliency map obtained by guided backpropagation \cite{springenberg2015striving}, and by GradCAM++ \cite{chattopadhay2018grad}.}
   \label{fig:gradcam}
\end{figure}

We present an example of saliency visualization using our model before and after adaptation on the target dataset in Figure \ref{fig:gradcam}. The results show that the model fails to localize pathological symptoms caused by potassium deficiency in crops without adapting the model to the target domain (Figure \ref{fig:gradcam}a). After adaptation, the model successfully classifies the nutrient status by focusing on the discriminative pathological symptoms (Figure \ref{fig:gradcam}b). 


\section{Conclusion}
\label{sec:conclusion}

In this work, we addressed unsupervised domain adaptation for plant nutrient deficiency detection. We proposed an approach that utilizes multiple views of crops both in the source and target domain. To this end, we enforce consistency between a query image and its strongly-augmented related view to incorporate complementary visual symptoms for the identification of plant nutrient deficiency. 
We also proposed a similarity-guided view mining mechanism, which ensures that the most dissimilar views are selected. We evaluated our approach on two datasets and provided additional ablation studies in the supplementary material. Our approach exhibited significantly improved performance compared to other methods in all the settings we studied. 
Despite the achieved major improvements compared to other unsupervised domain adaptation approaches, domain adaptation for nutrient deficiency detection across genotypes or cultivars remains challenging. Extending the approach to a multi-modal approach that utilizes data from other sensors, e.g., multi-spectral data, humidity, temperature, or chlorophyll fluorescence, is an interesting research direction to enhance the robustness and generalization abilities.

\paragraph{Acknowledgement} This work has been supported by the Deutsche Forschungsgemeinschaft (DFG, German Research Foundation) under Germany’s Excellence Strategy - EXC 2070–390732324 (PhenoRob), GA 1927/9-1 (KI-FOR 5351), and the Chinese Scholarship Council (202108440041).

\bibliography{egbib}

\newpage

\maketitle

\section{Supplementary Material}

\subsection{The MiPlo Datasets}
\label{sec:dataset}

The Mini Plot (MiPlo) datasets are large-scale RGB image datasets consisting of high-resolution images with multiple types of nutrient treatments annotated by agronomy experts. It consists of the Mini Plot Barley (MiPlo-B) dataset from Deichmann \etal\cite{deichmann2024rgb} and the newly collected Mini Plot Winter Wheat (MiPlo-WW) dataset. 

\paragraph{Experimental Setup}

The MiPlot experiments enable controlled trials of nutrient deficiencies under simulated field conditions. The crops (two genotypes) were grown in mineral soil in containers (“Big Box”, L x B x H: 111 x 71 x 61 cm, vol.: 535L) and sown in rows in a density according to agricultural practice. The soil was a mixture of nutrient-poor loess from a 5 meter depth of an opencast pit mine and quartz sand 0.5 – 1 mm. To expose the plants to environmental factors, \eg wind, radiation, and precipitation, the containers were positioned in an outdoor area and equipped with a fertigation system of porous drip pipes to allow additional water supply and individual fertilization with nutrient solutions per container.  To transfer soil microorganisms to the experiments, the containers were inoculated with soil slurry from the non-fertilized plot of a long-term fertilizer-field experiment. For each genotype, the containers were placed in three rows of ten containers each on a leveled concrete platform. The 30 containers were divided into seven treatments (ctrl, -N, -P, -K, -B, -S, unfertilized) with four replicates each, as well as two additional containers for invasive investigations, in a randomized block design. In this work, 24 containers with six nutrient treatments (ctrl, -N, -P, -K, -B, -S) were selected for evaluation, because the containers with unfertilized treatment showed distinct patterns, \ie only pipes. 


\paragraph{Image Acquisition Protocol}

The RGB images in the MiPlo datasets were taken from 24 containers, each of which was subjected to a type of nutrient treatment. 
All of the images with the size of 7296 $\times$ 5472 were captured by a Huawei P20 Pro smartphone with a triple camera from Leica under natural lighting conditions. 
Specifically, the images were taken under different conditions in terms of height, viewpoint, light, and weather to reflect realistic conditions. As a result, 
crops within each container have been captured over the growing season multiple times and each time from multiple views (20 views on average). Example images are shown in Figure \ref{fig:example}. The images were annotated by the date, genotype, and six nutrient treatments (ctrl, -N, -P, -K, -S, -B), where ``-'' stands for the omission of the corresponding nutrient (N: nitrogen, P: phosphorous, K: potassium, B: Boron, S:Sulfur). Plants in the control group `ctrl' do not suffer from nutrient deficiencies. 

\paragraph{Statistics}

The statistics of the MiPlo-B dataset \cite{deichmann2024rgb} and our proposed MiPlo-WW dataset are presented in Tables \ref{tab:statistics}, \ref{tab:statistics2}, and \ref{tab:statistics3}. \textbf{The Mini Plot Barley (MiPlo-B) dataset} consists of  18559 images with 6 nutrient treatments (-N, -P, -K, -B, -S, ctrl) annotated, ranging from 21.06.2022 - 20.07.2022 (16 dates). It contains two genotypes: Barke (9305 images) and Hanna (9254 images). For each genotype, each treatment was repeated 4 times, resulting in 24 containers, each of which has a unique ID. Six unique containers with six different nutrient treatments were selected as the test set while the other containers as the training set (\#train:\#test$\approx$75\%:25\%). 
\textbf{The Mini Plot Winter Wheat (MiPlo-WW) dataset} has 12466 images with 6 treatments (-N, -P, -K, -B, -S, ctrl) annotated, ranging from 12.05.2023 - 24.05.2023 (13 dates). It contains two genotypes: Julius (6253 images) and Meister (6213 images). The ID settings are the same as above.  
Although most annotations have a similar amount of images, there is a small imbalance of the sample distribution among different dates. 

\begin{table}[htbp!]
   \caption[Dataset statistics.]{Statistics of the MiPlo datasets. 
   }
      \begin{center}
         \begin{tabular}{@{}clllccccc@{}} 
            \toprule  
            \multicolumn{4}{c}{\textbf{Dataset}} & \textbf{\#Images (k)} & \textbf{\#Class} & \textbf{Dates} & \textbf{\#Views} & \textbf{Year} \\ 
            \midrule
            \multicolumn{4}{c}{MiPlo-B (Barley) \cite{deichmann2024rgb}} & 18.6 & 6 & 16 & 20 & 2022 \\	 
            \multicolumn{4}{c}{MiPlo-WW (Winter Wheat)} & 12.5 & 6 & 13 & 20  & 2023 \\	 
            \bottomrule
         \end{tabular}
      \end{center}
      \label{tab:statistics}
\end{table}

\begin{table}[htbp!] 
	\caption[Statistics of dataset]
	{The number of images in the Mini Plot Barley (MiPlo-B) dataset with two genotypes: Barke and Hanna. \textbf{06/21} denotes 21 June 2022, where 2022 is omitted for simplification. 
    \textbf{``-''} stands for the omission of the corresponding nutrient (N: nitrogen, P: phosphorous, K: potassium, B: Boron, S:Sulfur).
    }
	\begin{center}
		\setlength{\tabcolsep}{2pt}
            \resizebox{\textwidth}{!}{
		\begin{tabular}{@{}cccccccccccccccccccc|cc@{}} 
			\toprule  
			\multicolumn{4}{c}{\multirow{1}{*}{\textbf{No.}}}  &  \textbf{1}  &  \textbf{2}  &\textbf{3}  &\textbf{4}  &\textbf{5}  &\textbf{6}  &\textbf{7}  &\textbf{8}  &\textbf{9}  & \textbf{10} & \textbf{11} & \textbf{12} & \textbf{13} & \textbf{14} & \textbf{15} & \textbf{16} &  \\ 
			\multicolumn{4}{c}{\multirow{1}{*}{\textbf{class/date}}}  & \textbf{06/21}  & \textbf{06/22}  & \textbf{06/23}  & \textbf{06/24}  & \textbf{06/27}  & \textbf{06/29}  & \textbf{06/30}  & \textbf{07/01}  & \textbf{07/04} & \textbf{07/05} & \textbf{07/06} & \textbf{07/07} & \textbf{07/08} & \textbf{07/11} & \textbf{07/19} & \textbf{07/20}  &\textbf{Total}\\ 
            \hline
            \multicolumn{21}{c}{\textbf{Barke}} \\
                \hline
			\multicolumn{4}{c}{-N} & 80 & 84 & 80 & 80 & 161 & 160 & 78 & 80 & 80 & 80 & 156 & 80 & 80 & 168 & 60 & 99 & 1606 \\
			\multicolumn{4}{c}{-P} & 80 & 60 & 81 & 79 & 160 & 160 & 60 & 83 & 60 & 80 & 134 & 80 & 80 & 124 & 60 & 100 & 1481 \\	 
			\multicolumn{4}{c}{-K} & 80 & 80 & 80 & 80 & 163 & 158 & 80 & 80  & 60 & 80 & 152 & 80 & 80 & 148 & 39 & 99 & 1539 \\	
			\multicolumn{4}{c}{-B} & 81 & 80 & 81 & 80 & 164 & 140 & 83 & 80  & 80 & 81 & 137 & 80 & 80 & 130 & 58 & 100 & 1535 \\ 
		    \multicolumn{4}{c}{-S} & 80 & 82 & 80 & 80 & 161 & 160 & 83 & 80  & 80 & 80 & 135 & 80 & 80 & 138 & 59 & 100 & 1558 \\ 	
		    \multicolumn{4}{c}{ctrl} & 80 & 80 & 80 & 80 & 160 & 160 & 82 & 80 & 81 & 80 & 154 & 82 & 80 & 148 & 59 & 100 & 1586 \\	 
                \hline
			\multicolumn{4}{c}{total} & 481 & 466 & 482 & 479 & 969 & 938 & 466 & 483 & 441 & 481 & 868 & 482 & 480 & 856 & 335 & 598 & 9305 \\
                 \hline
                \multicolumn{21}{c}{\textbf{Hanna}} \\
                \hline
			\multicolumn{4}{c}{-N} & 80 & 80 & 80 & 80 & 160 & 140 & 82 & 79 & 80 & 80 & 154 & 80 & 80 & 144 & 59 & 99 & 1557 \\
			\multicolumn{4}{c}{-P} & 80 & 80 & 80 & 80 & 148 & 161 & 60 & 80 & 61 & 80 & 124 & 80 & 80 & 128 & 59 & 99 & 1481 \\	 
			\multicolumn{4}{c}{-K} & 80 & 80 & 80 & 80 & 160 & 160 & 81 & 80  & 60 & 80 & 152 & 80 & 80 & 154 & 39 & 100 & 1547 \\	
			\multicolumn{4}{c}{-B} & 83 & 82 & 81 & 80 & 161 & 140 & 81 & 81  & 80 & 81 & 137 & 80 & 80 & 134 & 59 & 101 & 1539 \\ 
		    \multicolumn{4}{c}{-S} & 80 & 80 & 81 & 80 & 162 & 162 & 81 & 80  & 81 & 80 & 134 & 80 & 80 & 144 & 60 & 100 & 1565 \\ 	
		    \multicolumn{4}{c}{ctrl} & 80 & 80 & 80 & 82 & 162 & 158 & 84 & 80 & 80 & 80 & 154 & 80 & 80 & 145 & 40 & 100 & 1565 \\	 	
                \hline
			\multicolumn{4}{c}{total} & 483 & 482 & 482 & 482 & 953 & 921 & 469 & 480 & 442 & 481 & 855 & 480 & 480 & 849 & 316 & 599 & 9254 \\
			\bottomrule
		\end{tabular}
            }
	\end{center}
	\label{tab:statistics2}
\end{table}

\begin{table}[htbp!] 
	\caption[Statistics of dataset]
	{The number of images in the Mini Plot Winter Wheat (MiPlo-WW) dataset with two genotypes: Julius and Meister. \textbf{05/12} denotes 12 May 2023, where 2023 is omitted for simplification. 
    \textbf{``-''} stands for the omission of the corresponding nutrient (N: nitrogen, P: phosphorous, K: potassium, B: Boron, S:Sulfur).
    }
	\begin{center}
		\setlength{\tabcolsep}{2pt}
            \resizebox{0.9\textwidth}{!}{
		\begin{tabular}{@{}ccccccccccccccccc|cc@{}} 
			\toprule  
   			\multicolumn{4}{c}{\multirow{1}{*}{\textbf{No.}}}  &  \textbf{1}  &  \textbf{2}  &\textbf{3}  &\textbf{4}  &\textbf{5}  &\textbf{6}  &\textbf{7}  &\textbf{8}  &\textbf{9}  & \textbf{10} & \textbf{11} & \textbf{12} & \textbf{13} &  \\ 
			\multicolumn{4}{c}{\multirow{1}{*}{\textbf{class/date}}}  & \textbf{05/12}  & \textbf{05/13}  & \textbf{05/14}  & \textbf{05/15}  & \textbf{05/16}  & \textbf{05/17}  & \textbf{05/18}  & \textbf{05/19}  & \textbf{05/20} & \textbf{05/21} & \textbf{05/22} & \textbf{05/23} & \textbf{05/24}  &\textbf{Total}\\ 
            \hline
            \multicolumn{18}{c}{\textbf{Julius}} \\
                \hline
			\multicolumn{4}{c}{-N} & 80 & 20 & 145 & 80 & 75 & 80 & 80 & 80 & 81 & 80 & 59 & 78 & 80 & 1018 \\
			\multicolumn{4}{c}{-P} & 80 & 61 & 79 & 80 & 80 & 82 & 83 & 81 & 80 & 80 & 81 & 81 & 83 & 1031 \\	 
			\multicolumn{4}{c}{-K} & 81 & 20 & 142 & 81 & 82 & 84 & 80 & 80  & 81 & 80 & 78 & 78 & 85 & 1052 \\	
			\multicolumn{4}{c}{-B} & 62 & 40 & 140 & 79 & 81 & 87 & 83 & 81  & 81 & 80 & 83 & 82 & 80 & 1059 \\ 
		    \multicolumn{4}{c}{-S} & 81 & 40 & 117 & 81 & 80 & 70 & 80 & 80  & 85 & 80 & 101 & 82 & 80 & 1057 \\ 	
		    \multicolumn{4}{c}{ctrl} & 81 & 20 & 143 & 80 & 83 & 62 & 84 & 80 & 80 & 80 & 81 & 81 & 81 & 1036 \\	 
                \hline
			\multicolumn{4}{c}{total} & 465 & 201 & 766 & 481 & 481 & 465 & 490 & 482 & 488 & 480 & 483 & 482 & 489 & 6253 \\
                 \hline
                \multicolumn{18}{c}{\textbf{Meister}} \\
                \hline
			\multicolumn{4}{c}{-N} & 80 & 20 & 140 & 78 & 80 & 80 & 82 & 83 & 86 & 81 & 77 & 79 & 73 & 1039 \\
			\multicolumn{4}{c}{-P} & 80 & 60 & 78 & 80 & 83 & 80 & 81 & 86 & 80 & 60 & 81 & 71 & 81 & 1001 \\	 
			\multicolumn{4}{c}{-K} & 80 & 20 & 144 & 87 & 86 & 82 & 83 & 81  & 81 & 80 & 80 & 75 & 80 & 1059 \\	
			\multicolumn{4}{c}{-B} & 78 & 40 & 150 & 81 & 81 & 82 & 83 & 80  & 80 & 80 & 81 & 78 & 74 & 1068 \\ 
		    \multicolumn{4}{c}{-S} & 81 & 40 & 120 & 80 & 80 & 82 & 83 & 80  & 81 & 80 & 78 & 79 & 75 & 1039 \\ 	
		    \multicolumn{4}{c}{ctrl} & 80 & 22 & 122 & 82 & 83 & 63 & 79 & 83 & 80 & 80 & 80 & 78 & 75 & 1007 \\	 
                \hline
			\multicolumn{4}{c}{total} & 479 & 202 & 754 & 488 & 493 & 469 & 491 & 493 & 488 & 461 & 477 & 460 & 458 & 6213 \\
			\bottomrule
		\end{tabular}
            }
	\end{center}
	\label{tab:statistics3}
\end{table}

\begin{figure}[p]   
	\centering
        \subfigure[MiPlo-B (Barke)]{ 
        \begin{minipage}[t]{1\linewidth}
            \centering
            \includegraphics[width=2cm, height=2cm]{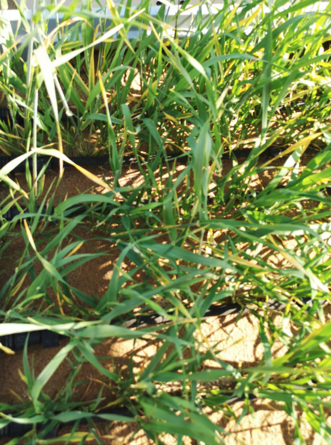}
            \includegraphics[width=2cm, height=2cm]{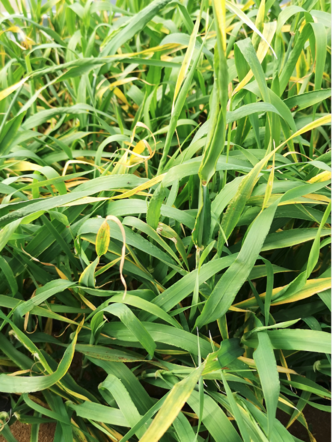}
            \includegraphics[width=2cm, height=2cm]{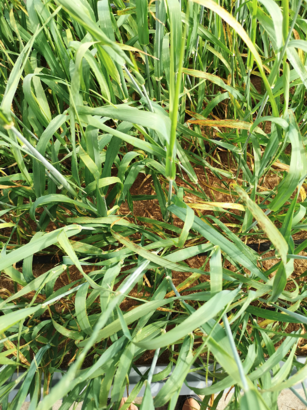}
            \includegraphics[width=2cm, height=2cm]{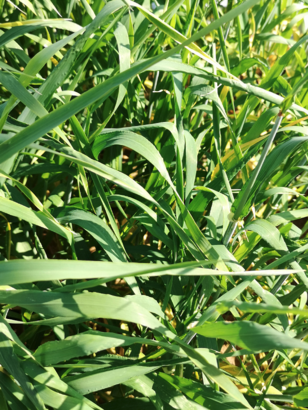}
            \includegraphics[width=2cm, height=2cm]{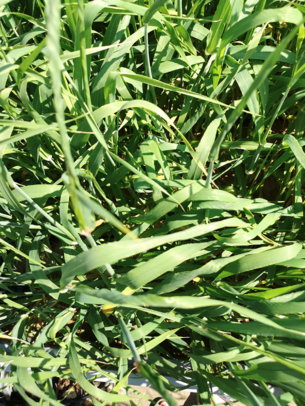}
            \includegraphics[width=2cm, height=2cm]{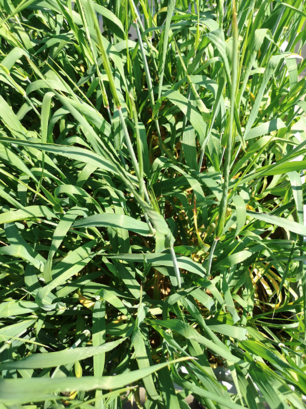} 
            \\
            \includegraphics[width=2cm, height=2cm]{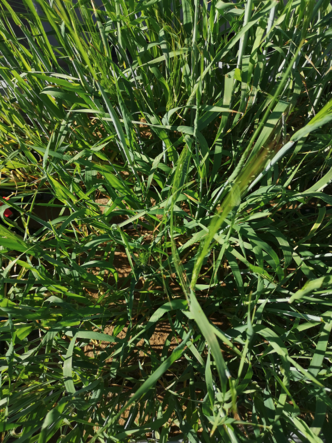}
            \includegraphics[width=2cm, height=2cm]{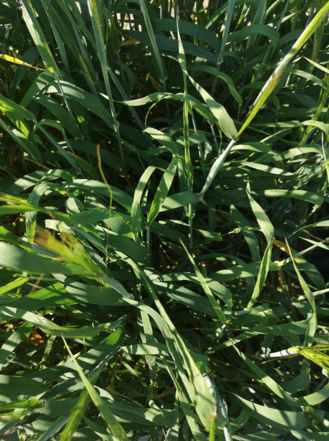}
            \includegraphics[width=2cm, height=2cm]{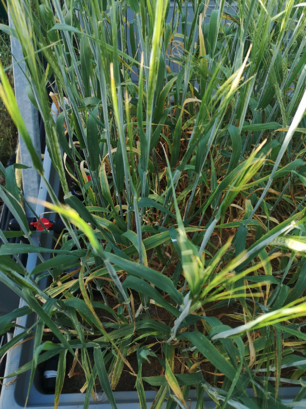}
            \includegraphics[width=2cm, height=2cm]{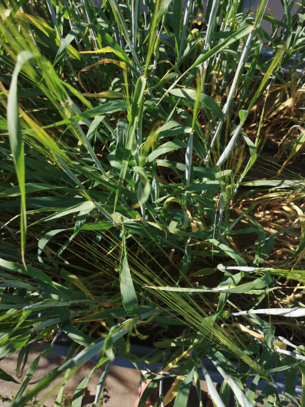}
            \includegraphics[width=2cm, height=2cm]{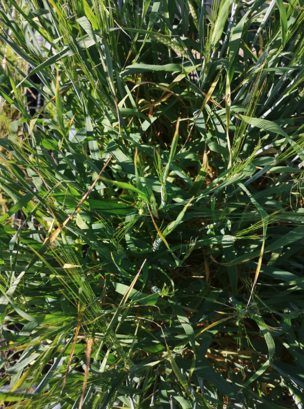}
            \includegraphics[width=2cm, height=2cm]{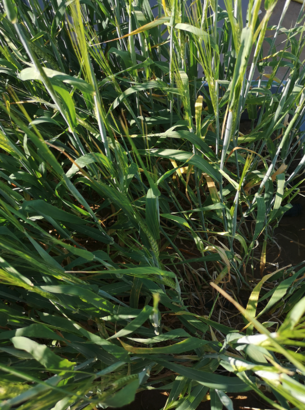}
            \\
            \includegraphics[width=2cm, height=2cm]{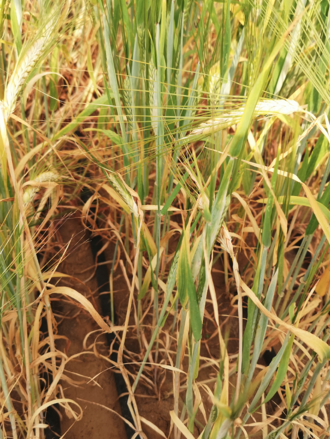}
            \includegraphics[width=2cm, height=2cm]{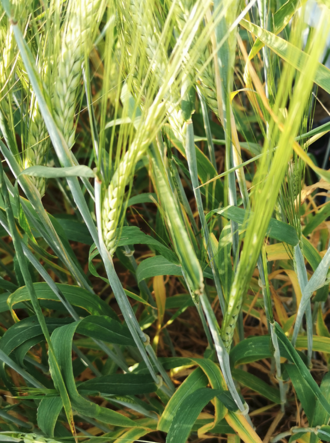}
            \includegraphics[width=2cm, height=2cm]{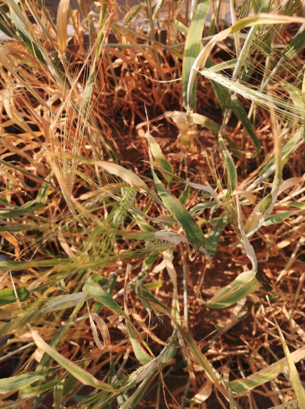}
            \includegraphics[width=2cm, height=2cm]{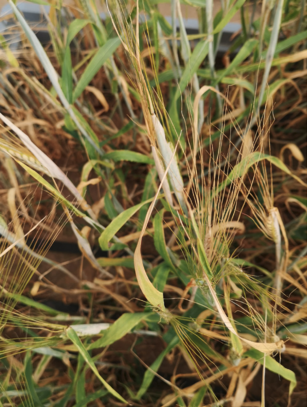}
            \includegraphics[width=2cm, height=2cm]{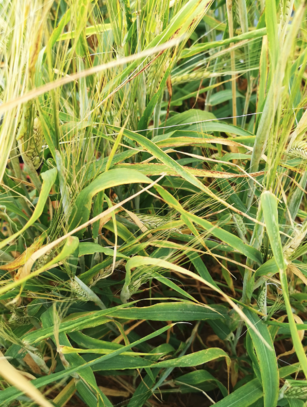}
            \includegraphics[width=2cm, height=2cm]{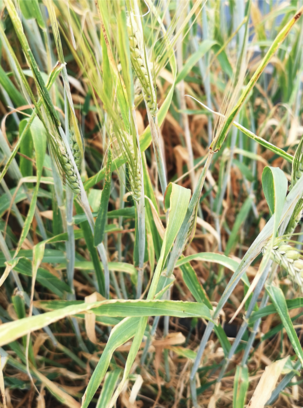}            
        \end{minipage}
        }
        \subfigure[MiPlo-B (Hanna)]{ 
        \begin{minipage}[t]{1\linewidth}
            \centering
            \includegraphics[width=2cm, height=2cm]{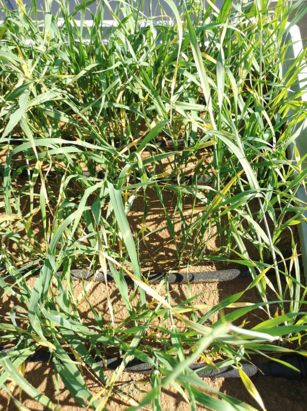}
            \includegraphics[width=2cm, height=2cm]{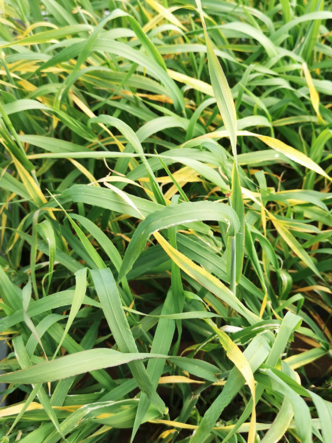} 
            \includegraphics[width=2cm, height=2cm]{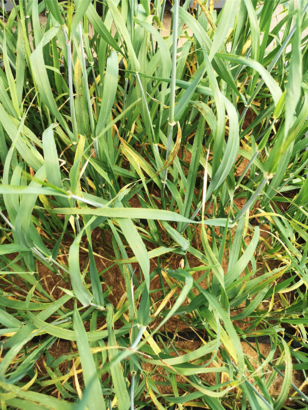}
            \includegraphics[width=2cm, height=2cm]{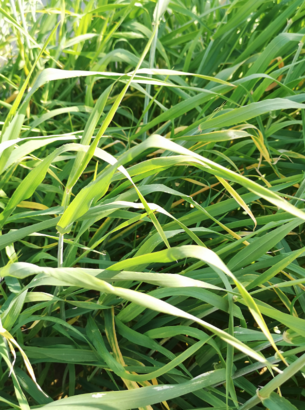}
            \includegraphics[width=2cm, height=2cm]{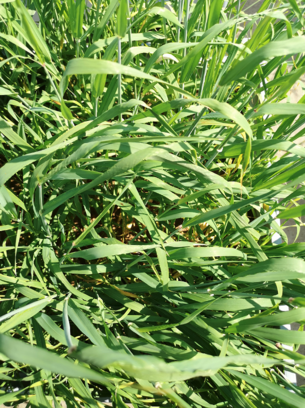}
            \includegraphics[width=2cm, height=2cm]{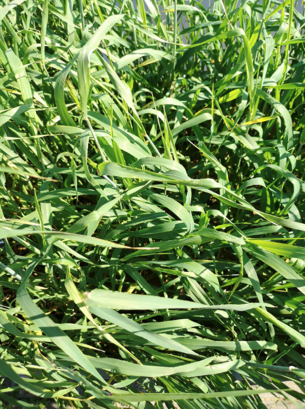}
            \\
            \includegraphics[width=2cm, height=2cm]{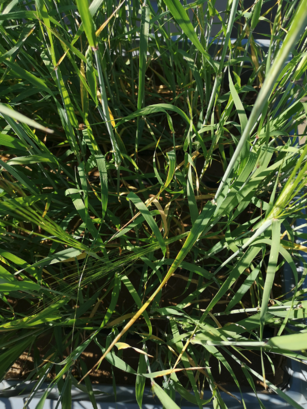}
            \includegraphics[width=2cm, height=2cm]{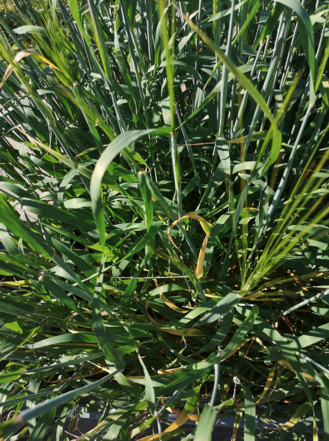}
            \includegraphics[width=2cm, height=2cm]{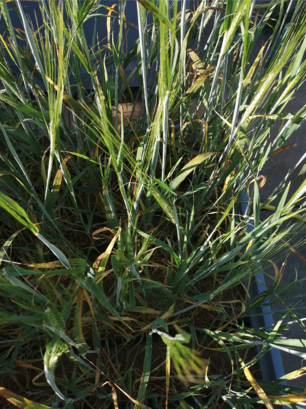}
            \includegraphics[width=2cm, height=2cm]{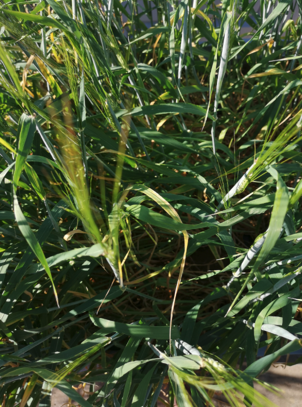}
            \includegraphics[width=2cm, height=2cm]{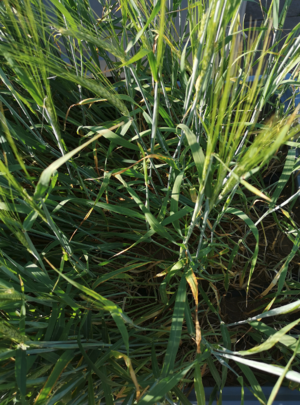}
            \includegraphics[width=2cm, height=2cm]{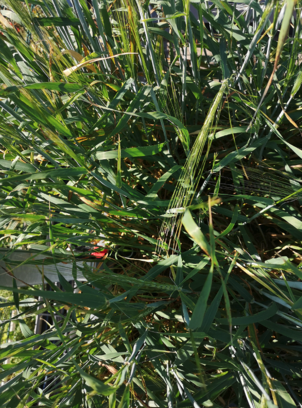}
            \\
            \includegraphics[width=2cm, height=2cm]{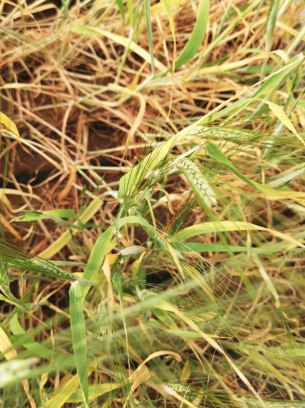}
            \includegraphics[width=2cm, height=2cm]{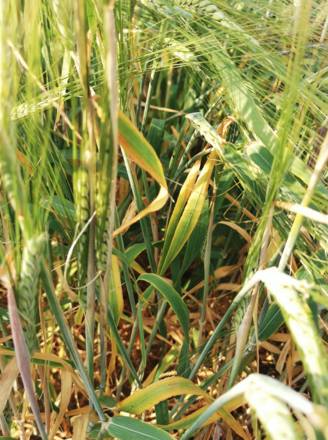}
            \includegraphics[width=2cm, height=2cm]{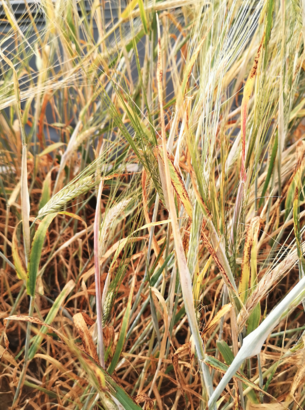}
            \includegraphics[width=2cm, height=2cm]{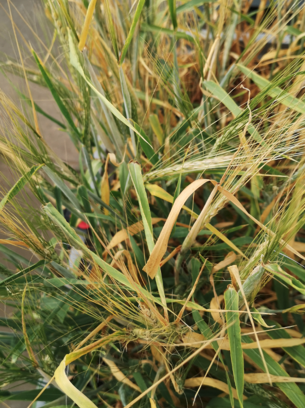}
            \includegraphics[width=2cm, height=2cm]{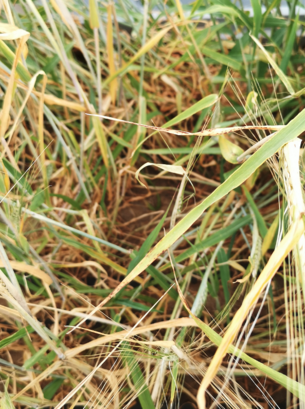}
            \includegraphics[width=2cm, height=2cm]{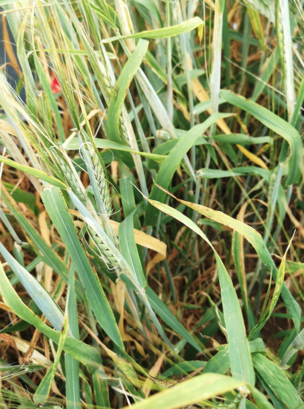}  
        \end{minipage}
        }

	\caption[Example images. ]
	{Example images. Columns 1-6: -N, -P, -K, -B, -S, ctrl; row 1-3: 21 June 2022, 04 July 2022, and 20 July 2022. 
 }
	\label{fig:example}
\end{figure} 

\begin{figure}[p]   
	\centering
        \subfigure[MiPlo-WW (Julius)]{ 
        \begin{minipage}[t]{1\linewidth}
            \centering
            \includegraphics[width=2cm, height=2cm]{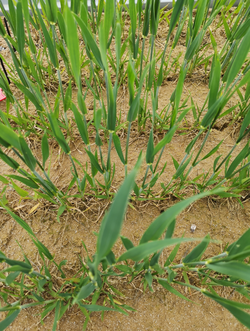}
            \includegraphics[width=2cm, height=2cm]{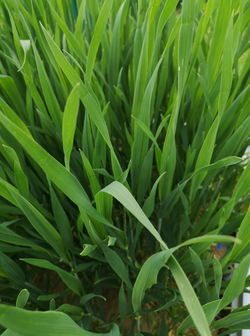}
            \includegraphics[width=2cm, height=2cm]{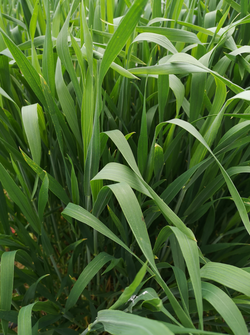}
            \includegraphics[width=2cm, height=2cm]{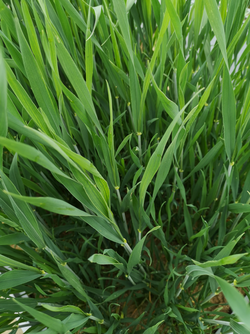}
            \includegraphics[width=2cm, height=2cm]{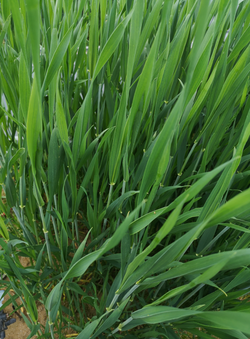}
            \includegraphics[width=2cm, height=2cm]{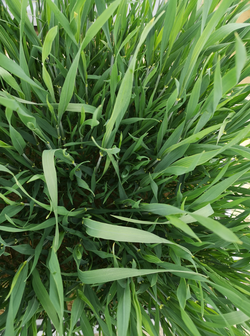} 
            \\
            \includegraphics[width=2cm, height=2cm]{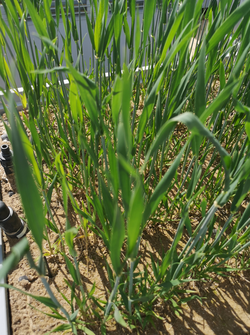}
            \includegraphics[width=2cm, height=2cm]{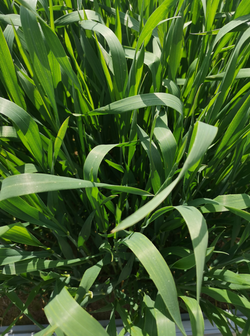}
            \includegraphics[width=2cm, height=2cm]{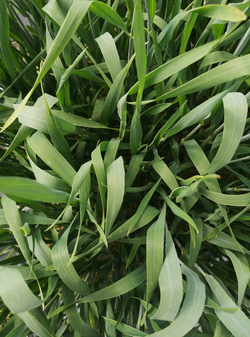}
            \includegraphics[width=2cm, height=2cm]{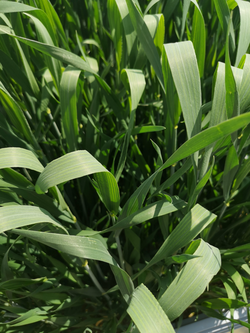}
            \includegraphics[width=2cm, height=2cm]{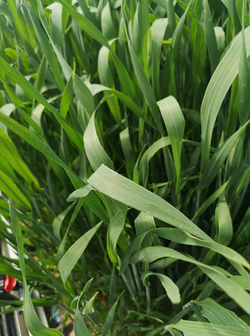}
            \includegraphics[width=2cm, height=2cm]{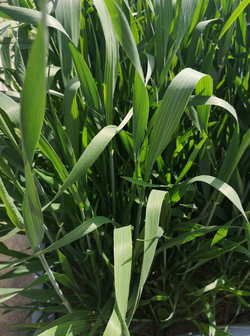}
            \\
            \includegraphics[width=2cm, height=2cm]{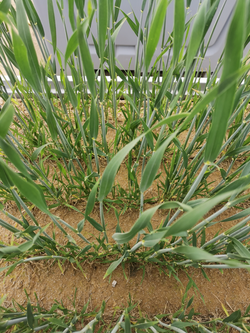}
            \includegraphics[width=2cm, height=2cm]{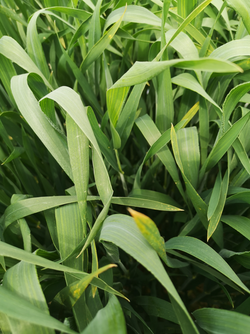}
            \includegraphics[width=2cm, height=2cm]{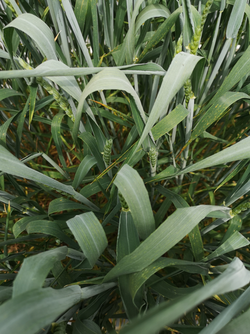}
            \includegraphics[width=2cm, height=2cm]{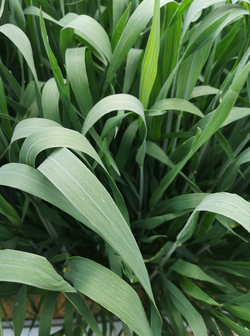}
            \includegraphics[width=2cm, height=2cm]{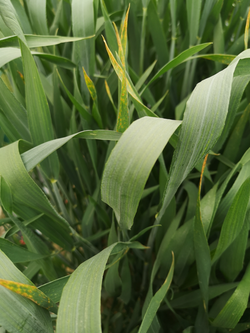}
            \includegraphics[width=2cm, height=2cm]{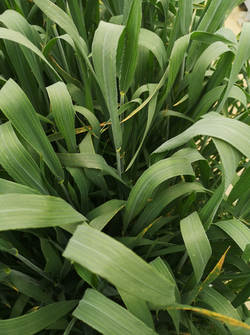}             
        \end{minipage}
        }
        \subfigure[MiPlo-WW (Meister)]{ 
        \begin{minipage}[t]{1\linewidth}
            \centering
            \includegraphics[width=2cm, height=2cm]{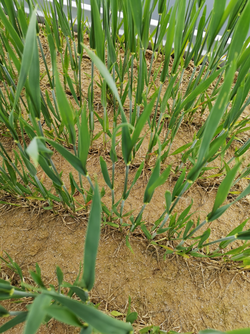}
            \includegraphics[width=2cm, height=2cm]{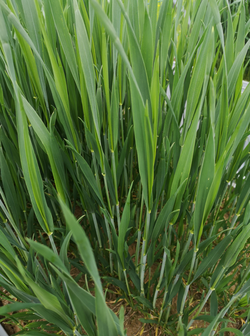} 
            \includegraphics[width=2cm, height=2cm]{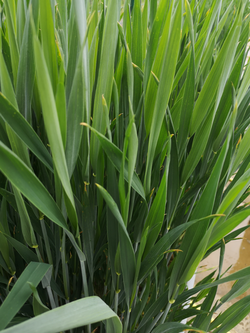}
            \includegraphics[width=2cm, height=2cm]{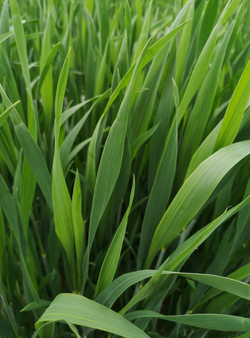}
            \includegraphics[width=2cm, height=2cm]{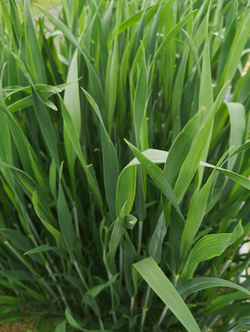}
            \includegraphics[width=2cm, height=2cm]{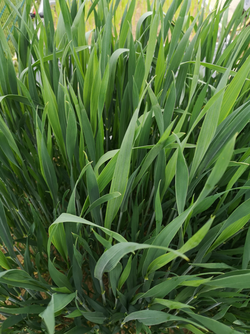}
            \\
            \includegraphics[width=2cm, height=2cm]{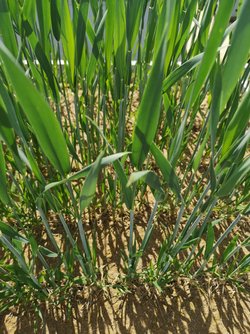}
            \includegraphics[width=2cm, height=2cm]{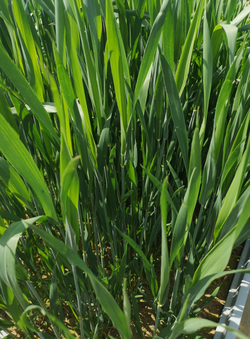}
            \includegraphics[width=2cm, height=2cm]{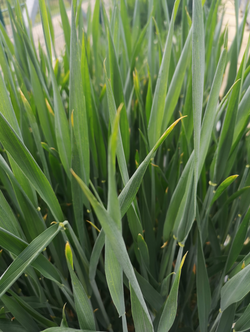}
            \includegraphics[width=2cm, height=2cm]{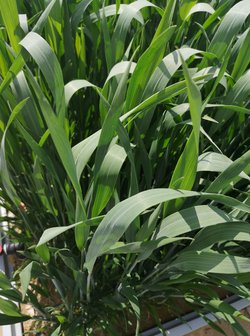}
            \includegraphics[width=2cm, height=2cm]{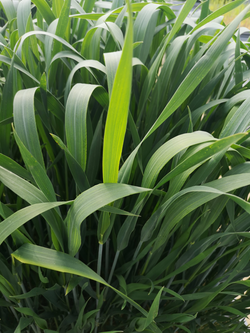}
            \includegraphics[width=2cm, height=2cm]{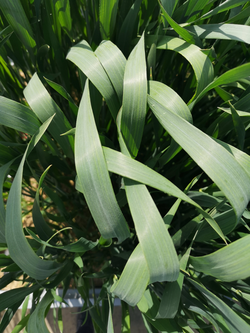}
            \\
            \includegraphics[width=2cm, height=2cm]{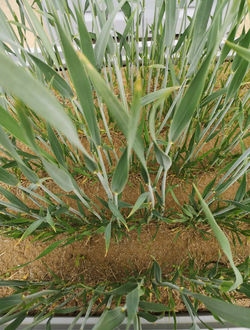}
            \includegraphics[width=2cm, height=2cm]{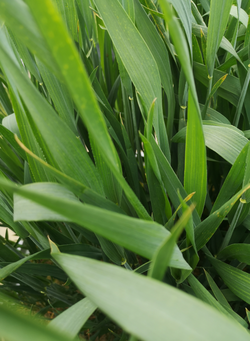}
            \includegraphics[width=2cm, height=2cm]{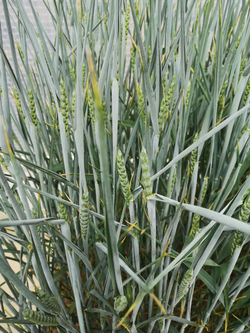}
            \includegraphics[width=2cm, height=2cm]{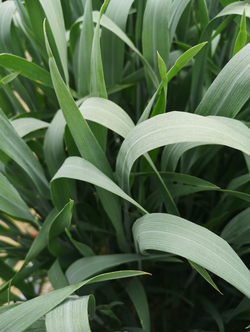}
            \includegraphics[width=2cm, height=2cm]{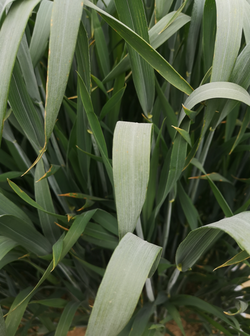}
            \includegraphics[width=2cm, height=2cm]{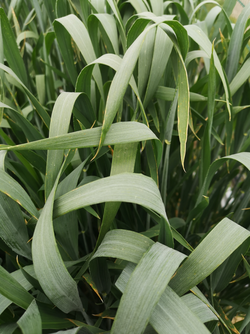}   
        \end{minipage}
        }
	\caption[Example images. ]
	{Example images. Column 1-6: -N, -P, -K, -B, -S, ctrl; 
 row 1-3: 12 May 2023, 18 May 2023, and 24 May 2023. 
 }
	\label{fig:example}
\end{figure} 

\subsection{Experimental Details}

If not specifically pointed out, the default setting of our model adopts the loss term $L_s^{sa2}$, $L_t^{sa2}$ and $L_t^{sa1}$, five subsampled views as well as a threshold of 0.8 for hard pseudo-labels for the ablation studies. 
And the default hyper-parameters in our experiments are as follows: 
The original image was resized to 1344 $\times$ 1344 and normalized with a mean value of [0.485, 0.456, 0.406] and a standard deviation of [0.229, 0.224, 0.225] calculated from ImageNet. 
For weak augmentation, we apply random horizontal flip with a probability of 50\%. 
Following previous work \cite{sohn2020fixmatch}, we adopted RandAugment, a variant of AutoAugment that does not need to pre-train the augmentation strategy with labeled data, for strong augmentation. 
We used ResNet-50 as backbone that was pre-trained on ImageNet. 
We then trained each model for 20 epochs with a batch of four samples from the source domain and four samples from the target domain at each iteration. We used stochastic gradient descent (SGD) with an initial learning rate of $3\times10^{-3}$, where the momentum and weight decay were set as 0.9 and $10^{-3}$, respectively. 
The learning rate was reduced with schedule ${lr}_p = \frac{{lr}_0}{(1+\alpha p)^{\beta}}$ where $p$ is the training progress linearly changing from 0 to 1, ${lr}_0$ is the initial learning rate, $\alpha=8$ and $\beta=0.75$ is the decay factor. 
All of the experiments were conducted with a single NVIDIA RTX A6000 with 48GB VRAM. 
For evaluation, we report the top-1 accuracy metric on the test set of the target domain, which denotes whether the predicted category with the highest confidence matches the ground truth category. 

\subsection{Additional Ablation Study}

\subsubsection{Number of Views}

To explore how many views are necessary for adaptation in the detection of plant nutrient deficiency, we sample a subset of related views by computing the similarity of each query-view pair given a query image and select the top $n$ related views that are most dissimilar to the query image. 
To validate the effectiveness of our proposed SgVM mechanism, we also report the results by randomly sampling the subset. The results in Table \ref{tab:view} indicate that the model with a subset of five related views performs the best, and including more views does not improve the performance but might provide noisy signals due to redundant information. Notably, model sampling with our proposed SgVM mechanism consistently outperforms its counterpart with random sampling. 

\vspace{.4cm}
\begin{minipage}{0.45\textwidth}
   \begin{center}
   \begin{tabular}{|c|c|c|}
      \hline
      \#Views & SgVM & Random \\ 
      \hline
      1 & 65.1 & 62.5 \\
      5 & \textbf{69.6} & \textbf{65.3} \\
      10 & 67.4 & 64.7 \\
      20 & 66.8 & 63.3 \\
      40 & 69.1 & 64.8 \\
      \hline
   \end{tabular}
   \end{center}
   \captionof{table}{Ablation study on the number of views and sampling methods for the Barley: B $\rightarrow$ H benchmark. \textit{\#Views} indicates the number of related views given a query image, \textit{SgVM} denotes our proposed \textbf{S}imilarity-\textbf{g}uided \textbf{V}iew \textbf{M}ining mechanism, and \textit{Random} refers to random sampling while constructing the set of views for each query image.}
   \label{tab:view}
\end{minipage}
\hspace{4pt}
\begin{minipage}[c]{0.45\textwidth}
   \begin{center}
   \begin{tabular}{|l|c|c|}
      \hline
      $\tau$ & B $\rightarrow$ H & B $\rightarrow$ M \\ 
      \hline
      0.3  & 66.0 & 34.0 \\
      0.5  & 63.0 & 31.2 \\
      0.8  & \textbf{69.6} & 33.7  \\ 
      0.9  & 64.4 & 31.2 \\
      \hline
      soft & 67.4 & \textbf{45.6} \\ 
      \hline
   \end{tabular}
   \end{center}
   \captionof{table}{Ablation study on threshold $\tau$ and types of pseudo-labels. B $\rightarrow$ H indicates smaller domain gap (cross genotypes), while B $\rightarrow$ M indicates larger domain gap (cross cultivars). \textit{soft} refers to soft pseudo-labels instead of hard pseudo-labels with a pre-defined threshold $\tau$.}
   \label{tab:softhard}
   \vspace{0.4cm}
\end{minipage}

\subsubsection{Hard Pseudo-Label vs.\ Soft Pseudo-Label}

To evaluate the impact of the threshold $\tau$ for hard pseudo-labels, as well as to compare hard pseudo-labels with soft pseudo-labels, we report the results in Table \ref{tab:softhard}. Comparing B $\rightarrow$ H and B $\rightarrow$ M adaptation, we see soft pseudo-labels work better than hard pseudo-labels when the domain gap is large (B $\rightarrow$ M), \ie the adaptation across crop species. In this case, increasing the threshold will mask out most of the pseudo-labels since the confidences are in general low, while decreasing the threshold will force the model to learn from noisy pseudo-labels. If the domain gap is smaller (B $\rightarrow$ H), the initial confidences are higher and the hard pseudo-labels perform better. The choice of hard and soft pseudo-labels thus depends on the domain gap, but soft pseudo-labels can always be applied.          


\subsubsection{Supervision of $\mathcal{L}_s^{sa2}$}
While we compute $\mathcal{L}_s^{gt}$ based on the ground-truth labels of the source images, $\mathcal{L}_s^{sa2}$ is computed based on the pseudo-labels. In Table \ref{tab:label}, we compare the results of computing the consistency loss $\mathcal{L}_s^{sa2}$ for source domain images with ground-truth labels or pseudo-labels. The results show that it is better to use the pseudo-labels instead of the ground-truth labels for each view pair of the query source images, which shows that the gain of $\mathcal{L}_s^{sa2}$ is due to measuring the prediction consistency between two views and not simply due to data augmentation.      

\begin{table}[ht]
   \begin{center}
   \begin{tabular}{|c|c|c|}
      \hline
      Supervision & B $\rightarrow$ H  & B $\rightarrow$ M \\ 
      \hline
      label & 66.8  & 35.9 \\ 
      pseudo-label & 69.6 (hard)  & 45.6 (soft) \\
      \hline
   \end{tabular}
   \end{center}
   \captionof{table}{Ablation study on the supervision signals of $\mathcal{L}_s^{sa2}$. }
   \label{tab:label}
\end{table} 
\begin{table}[H]
\end{table}

\begin{table}[t] 
   \caption[SOTA B2H]
      {Top-1 Classification Accuracy (\%) for adaptation across genotypes: \textbf{Barley: Barke $\rightarrow$ Hanna} with different backbones. 
      \textit{FLOPs} stands for floating point operations. 
      \textit{Oracle} indicates the model was trained with full supervision on the \textbf{Hanna} training set. \textit{Source-Only} denotes the results without adaptation. 
      The highest accuracy is shown in bold, while the second best is underlined.  
      }
      \begin{center}
         \resizebox{0.8\textwidth}{!}{
         \begin{tabular}{cccc|c|c|c} 
            \toprule  
            \multicolumn{4}{c}{\multirow{2}{*}{\textbf{Model}}}  &  \multicolumn{3}{c}{\textbf{Barley: Barke $\rightarrow$ Hanna}} \\ 
            \cmidrule{5-7}
            \multicolumn{4}{c}{} & \textbf{ResNet50}  &  \textbf{MobileNetV3Large} &  \textbf{MobileNetV3Small} \\  
            \hline
                  \multicolumn{4}{c}{Parameters (M)} & 25.6 & 5.5 & 2.5 \\ 
                  \multicolumn{4}{c}{FLOPs (G)} & 4.09 & 0.22 & 0.06 \\ 
                  \hline
                  \multicolumn{4}{c}{Oracle} & 75.4 & 70.2 & 67.7 \\ 
                  \hline
                  \multicolumn{4}{c}{Source-Only} & 54.0  & 53.0 & 48.0 \\ 
                   \multicolumn{4}{c}{DANN \cite{ganin2016domain}} & 54.7 & 47.0 & 41.0 \\ 
                   \multicolumn{4}{c}{ADDA \cite{tzeng2017adversarial}} & 51.8 & 39.5 & 27.8 \\ 
                   \multicolumn{4}{c}{JAN \cite{long2017deep}} &  51.9  & 49.4 & 43.2 \\ 
                   \multicolumn{4}{c}{CDAN \cite{long2018conditional}}  & 51.3 & 46.3 & 37.1  \\ 
                   \multicolumn{4}{c}{BSP \cite{chen2019transferability}} & 58.8 & 49.3 & 44.6 \\ 
                   \multicolumn{4}{c}{AFN \cite{xu2019larger}} & 59.0 & 51.3 & 45.1 \\ 
                   \multicolumn{4}{c}{Mean Teacher \cite{tarvainen2017mean}} & 60.6 & 56.2 & 48.7 \\ 
                   \multicolumn{4}{c}{FixMatch \cite{sohn2020fixmatch}} & 64.6 & \underline{57.5} & 49.6 \\ 
                  \multicolumn{4}{c}{FlexMatch \cite{zhang2021flexmatch}} & 65.9 & 57.1 & \underline{50.1}  \\ 
                  \multicolumn{4}{c}{Ours+hard} & \textbf{69.6} & \textbf{59.6} & \textbf{53.1} \\ 
                   \multicolumn{4}{c}{Ours+soft} & \underline{67.4} & 55.4 & 45.6 \\ 
                   \midrule         
         \end{tabular} }
      \end{center}
      \label{tab:B2H_backbone}
\end{table}

\subsubsection{Different Backbones} 

To explore the performance of smaller backbones with fewer parameters, as they are commonly used for applications with very limited computational resources, we also report the results for MobileNetV3 (large and small versions) \cite{howard2019searching} as backbone. The results in Table \ref{tab:B2H_backbone} show that our approach outperforms other baselines consistently with ResNet50 \cite{he2016deep} as well as various efficient MobileNetV3 architectures as backbones. 

\subsubsection{Confusion Matrices}
\begin{figure}[t]
\centering
\subfigure[B $\rightarrow$ H (without adaptation), AVG=54.0\%]{
\includegraphics[draft=false,trim={0 0 6cm 2.7cm},clip,width= 0.48\textwidth]{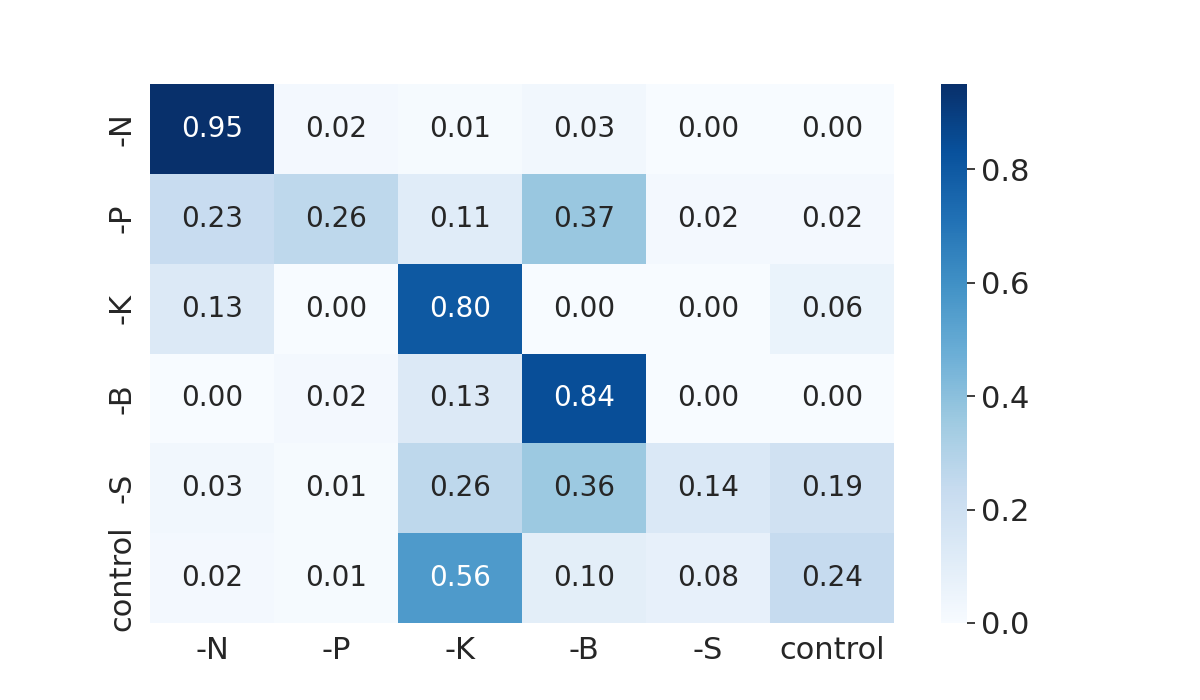}}
\subfigure[B $\rightarrow$ H (MV-Match+hard), AVG=69.6\%]{
\includegraphics[draft=false,trim={0 0 6cm 2.7cm},clip,width= 0.48\textwidth]{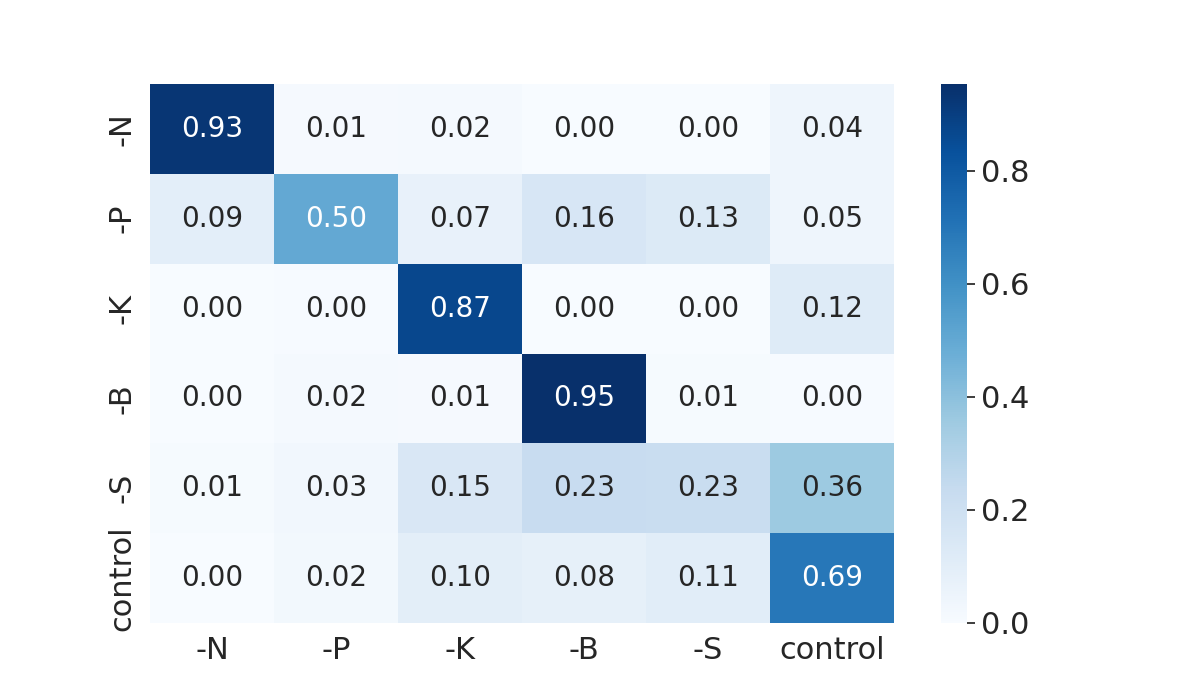}}
\subfigure[B $\rightarrow$ J (without adaptation). AVG=31.9\%]{
\includegraphics[draft=false,trim={0 0 6cm 2.7cm},clip,width= 0.48\textwidth]{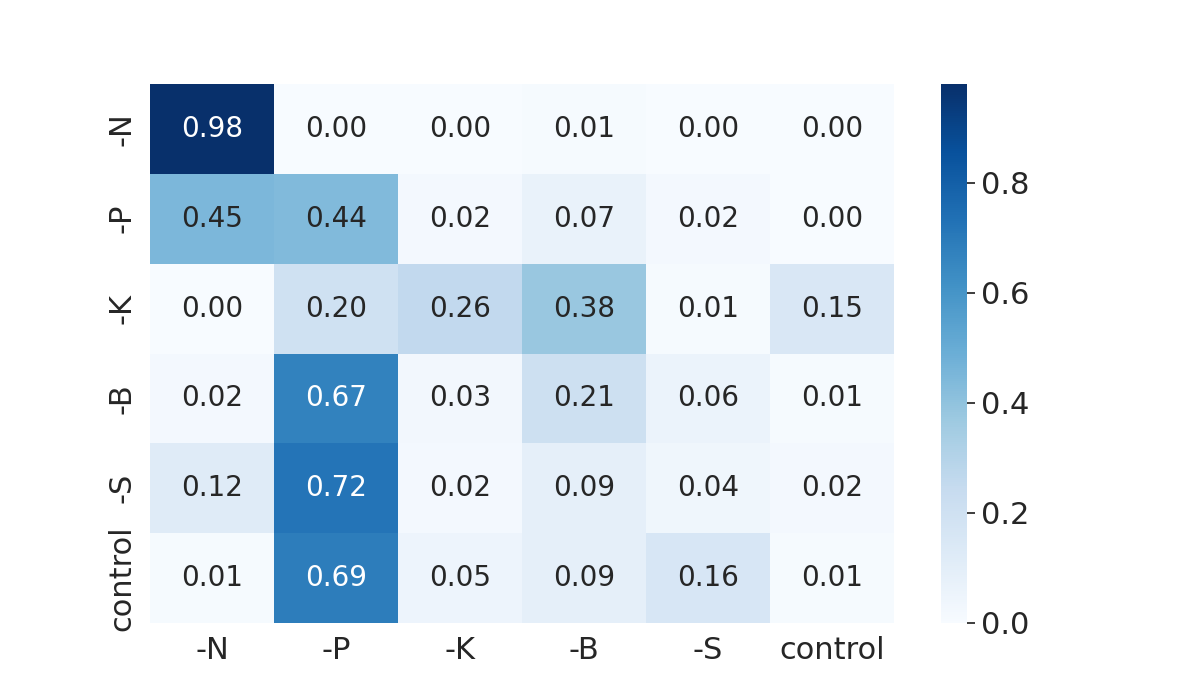}}
\subfigure[B $\rightarrow$ J (MV-Match+soft). AVG=42.2\%]{
\includegraphics[draft=false,trim={0 0 6cm 2.7cm},clip,width= 0.48\textwidth]{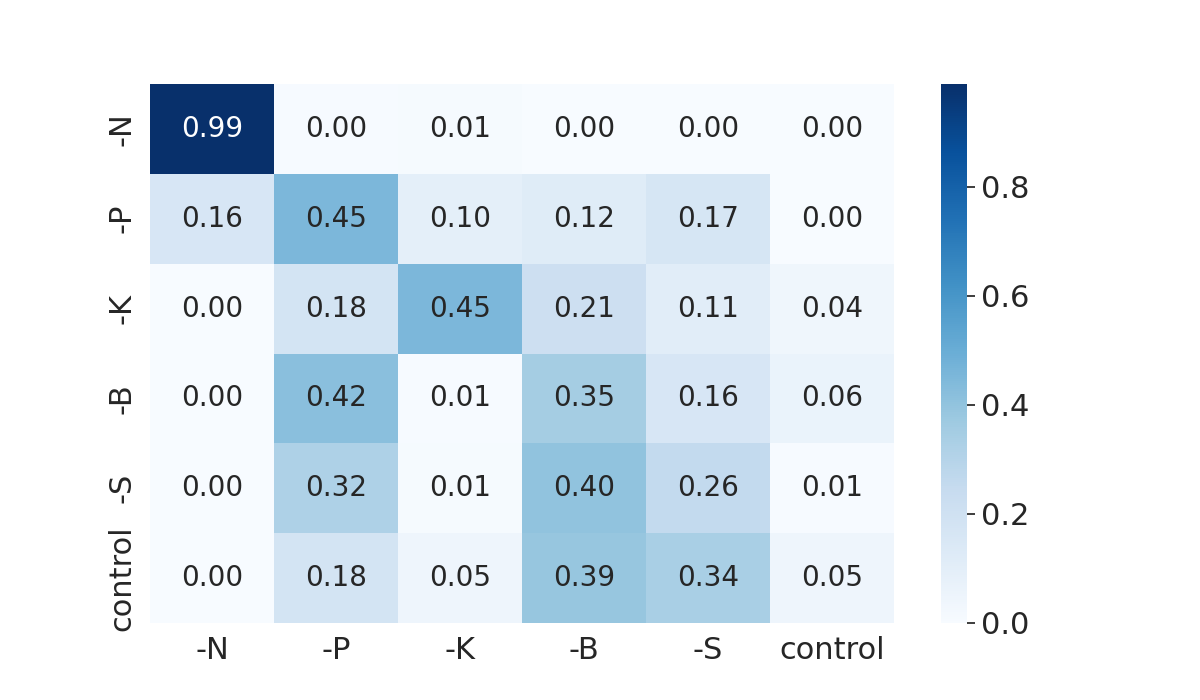}}
\caption{Confusion matrices (x-axis: predicted treatment, y-axis: real treatment) without adaptation (a) and our approach (b) for \textbf{Barley: Barke $\rightarrow$ Hanna}. The confusion matrices (c,d) are without adaptation and our approach for \textbf{Barley $\rightarrow$ Winter Wheat (B $\rightarrow$ J)}. \textit{AVG} denotes average accuracy.  } 
\label{fig:confmat} 
\end{figure}

We finally show confusion matrices before and after adaptation in Figure \ref{fig:confmat}. 

\end{document}